\documentclass[sigconf]{acmart}

\setcopyright{acmlicensed}
\usepackage{amsmath}
\usepackage{amsfonts}
\usepackage{graphicx}
\usepackage{balance}
\usepackage{subcaption}
\usepackage{booktabs} 
\usepackage{enumitem}
\usepackage{tabularx}
\usepackage{multirow}
\newtheorem{theorem}{Theorem}

\newtheorem{definition}{Definition}

\newtheorem{assumption}{Assumption}
\usepackage{hyperref}

\copyrightyear{2026}
\acmYear{2026}
\setcopyright{cc}
\setcctype{by}
\acmConference[KDD '26]{Proceedings of the 32nd ACM SIGKDD Conference on Knowledge Discovery and Data Mining V.2}{August 09--13, 2026}{Jeju Island, Republic of Korea}
\acmBooktitle{Proceedings of the 32nd ACM SIGKDD Conference on Knowledge Discovery and Data Mining V.2 (KDD '26), August 09--13, 2026, Jeju Island, Republic of Korea}
\acmDOI{10.1145/3770855.3817724}
\acmISBN{979-8-4007-2259-2/2026/08}

\begin{document}

\title{Beyond Semantic Understanding: Preserving Collaborative Frequency Components in LLM-based Recommendation}

\author{Minhao Wang}
\affiliation{%
  \institution{East China Normal University}
  \city{Shanghai}
  \country{China}
}
\email{51275901104@stu.ecnu.edu.cn}

\author{Yunhang He, Cong Xu}
\affiliation{%
  \institution{East China Normal University}
  \city{Shanghai}
  \country{China}
}
\email{ yhhe2004, congxueric@gmail.com}



\author{Zhangchi Zhu}
\affiliation{%
  \institution{East China Normal University}
  \city{Shanghai}
  \country{China}
}
\email{zczhu@stu.ecnu.edu.cn}



\author{Shuang Hao}
\affiliation{%
  \institution{Beijing Jiaotong University}
  \city{Beijing}
  \country{China}
}
\email{haoshuang@bjtu.edu.cn}

\author{Ning Liu}
\affiliation{%
  \institution{Shandong University}
  \city{Jinan, Shandong}
  \country{China}
}
\email{victorliucs@gmail.com}

\author{Wei Zhang}
\authornote{Corresponding author. This work was supported in part by National Key R\&D Program of China (2023YFC3341200), National Natural Science Foundation of China (62572198, 62372034, and 62402294), and the Key Laboratory of Advanced Theory and Application in Statistics and Data Science, Ministry of Education.}
\affiliation{%
  \institution{East China Normal University \& Shanghai Innovation Institute}
  \city{Shanghai}
  \country{China}
}
\email{zhangwei.thu2011@gmail.com}

\begin{abstract}
Recommender systems in concert with Large Language Models (LLMs) present promising avenues for generating semantically-informed recommendations. However, LLM-based recommenders exhibit a tendency to overemphasize semantic correlations within users' interaction history. When taking pretrained collaborative ID embeddings as input, LLM-based recommenders progressively weaken collaborative signals layer by layer, contrary to traditional Transformer-based sequential models where such signals are typically preserved or even enhanced. To address this limitation, we introduce FreLLM4Rec, an approach designed to balance semantic and collaborative information from a spectral perspective. Item embeddings that incorporate both semantic and collaborative information are first purified using a Global Graph Low-Pass Filter (G-LPF) to preliminarily remove irrelevant high-frequency noise. Temporal Frequency Modulation (TFM) then actively preserves collaborative signal layer by layer. Note that the collaborative preservation capability of TFM is theoretically guaranteed by establishing a connection between the optimal but hard-to-implement local graph fourier filters and the suboptimal yet computationally efficient frequency-domain filters. Extensive experiments on four benchmark datasets demonstrate that FreLLM4Rec successfully mitigates collaborative signal attenuation and achieves competitive performance, with improvements of up to 8.00\% in NDCG@10 over the best baseline. Our findings provide insights into how LLMs process collaborative information and offer a principled approach for improving LLM-based recommendation systems. Our code is available at \url{https://github.com/BDML-lab/FreLLM4Rec}.
\end{abstract}



\maketitle

\section{Introduction}

Large Language Models, when used as recommenders, offer a novel perspective compared to traditional models that primarily focus on fitting user-item interaction data ~\cite{fan2023recommender, wu2023survey, liu2023pre}. Recent advances have demonstrated that LLMs can effectively leverage their pre-trained knowledge to address long-standing challenges in recommendation systems, including cold-start problems, cross-domain transfer, and explainability~\cite{zhao2023survey, zhai2024actions, wang2025oagir}. 
LLM-based recommendation has evolved from pure text-based methods~\cite{geng2022recommendation, bao2023tallrec} to hybrid approaches combining semantic understanding with collaborative signals~\cite{wang2025oagir, yuan2025cilo, liu2025tmf, chen2025tler, he2025llm2reclargelanguagemodels}, as semantics alone cannot capture the collaborative patterns essential for recommendation~\cite{liao2024llara, wei2024llmrec, ren2025llmcf}. The most effective systems successfully balance both.

\begin{figure}[!t]
 \centering
 \begin{subfigure}[b]{0.95\linewidth}
  \centering
  \includegraphics[width=\textwidth]{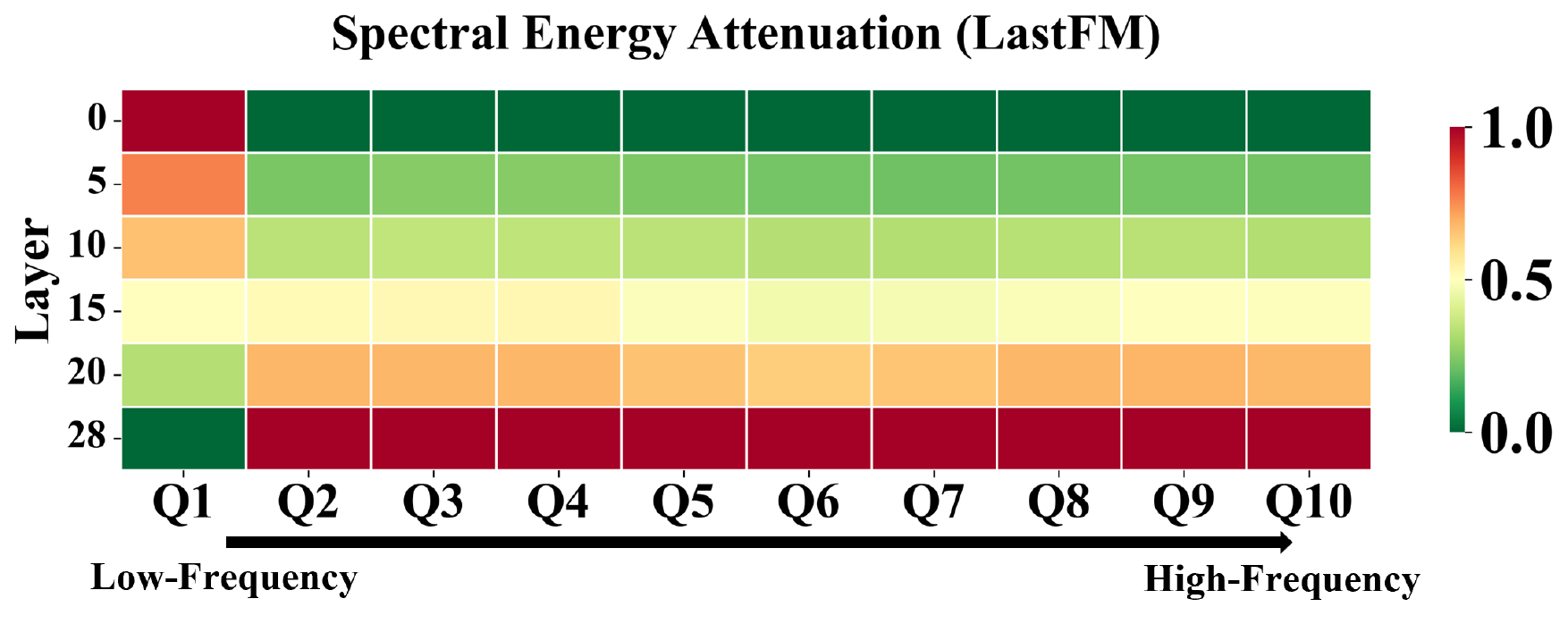}
  \vspace{-2.em}
  \caption{Vanilla LLM}
  \label{fig:decay_phenomenon_detailed}

 \end{subfigure}
 
 \begin{subfigure}[b]{0.95\linewidth}
  \centering
  \includegraphics[width=\textwidth]{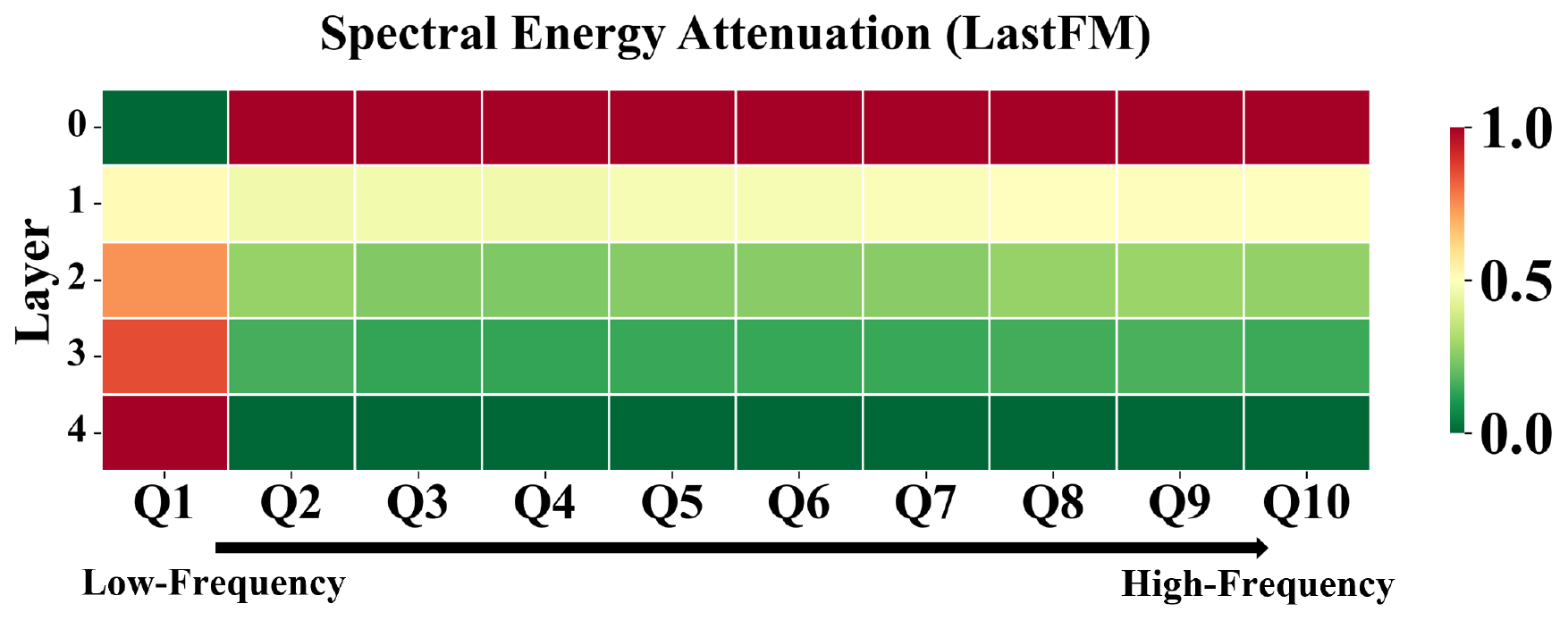}
    \vspace{-2.em}
  \caption{SASRec}
  \label{fig:sasrec_preserved}
 \end{subfigure}
 
\caption{Discovery of Intra-Layer Spectral Attenuation. X-axis: frequency bands (Q1=low to Q10=high); Y-axis: layer depth. Color intensity shows energy. Note these heatmaps apply frequency-wise (column-wise) min-max normalization to highlight relative cross-layer trends for each frequency component, making the evolution patterns more visible. (a) LLMs progressively lose low-frequency collaborative energy across layers. (b) SASRec preserves these signals, proving attenuation is LLM-specific, not architecture-inherent.}
 \label{fig:comparison_decay}
\end{figure}

Despite these advances, existing approaches predominantly treat LLMs as black boxes, focusing on input-output relationships without investigating how collaborative information evolves within the model's internal mechanisms. This oversight has led to a fundamental gap in our understanding: how do LLMs' deep architectures process and transform the collaborative signals that are essential for recommendation? This paper addresses this critical gap by conducting the first systematic investigation of collaborative signal transformation within LLM architectures from a spectral perspective.
To be specific, 
recent developments in graph spectral theory have shown that collaborative information primarily lies in the low-frequency components~\cite{wu2021self, yu2023xsimgcl, he2020lightgcn, zhu2025exploring} derived from a local interaction graph.
If the sequence embeddings are predominantly characterized by low-frequency components, they are assumed to retain the majority of collaborative information; conversely, a dominance of high frequencies will suggest the aforementioned attenuation~(see \figurename~\ref{fig:decay_phenomenon_detailed}).
In contrast, SASRec~\cite{kang2018self, vaswani2017attention} preserves and even enhances low-frequency components (Figure~\ref{fig:sasrec_preserved}), with consistent observations across backbones and datasets (Figure~\ref{fig:other_dataset_decay}).
We term this phenomenon \textit{Intra-Layer Spectral Attenuation}, which arises because LLM-based recommenders overly depend on their internal knowledge and reasoning abilities
while implicitly resisting the integration of collaborative information.

A direct and theoretically optimal solution is to apply low-pass filtering based on the corresponding local graphs~\cite{shuman2013emerging, ortega2018graph} 
to retain the majority of collaborative information.
Yet two practical barriers arise: 
(\romannumeral1) sampling local graphs is intricate and poorly parallelizable, making the method prohibitively expensive~\cite{defferrard2016convolutional}, and (\romannumeral2) the approach ignores temporal dynamics, discarding sequential patterns that capture evolving user preferences.
Fortunately, under certain mild assumptions, we prove that equivalent collaborative information can be retained by temporal frequency-domain filtering~\cite{zhou2022filter}. Specifically, by transforming the signals via the computationally efficient Discrete Fourier Transform (DFT) and attenuating high-frequency components, we achieve a theoretically analogous effect to graph spectral filtering while avoiding expensive eigendecomposition.
Although frequency analysis has been applied in recommendation systems~\cite{du2023frequency, zhou2022filter, shin2024attentiveinductivebiassequential} and LLM fine-tuning~\cite{gao2024parameter, zhao2024spectral}, our work differs fundamentally in both motivation and methodology. Prior work uses Fourier transforms as architectural components for model design, whereas we employ Graph Signal Processing as a diagnostic tool to systematically investigate how collaborative signals evolve within pre-trained LLM architectures. This spectral analysis reveals the previously unidentified phenomenon of Intra-Layer Spectral Attenuation. Our use of temporal frequency modulation then serves as a corrective mechanism grounded in this diagnostic finding, with Theorem~\ref{thm:temporal_to_graph} establishing the theoretical connection between efficient temporal filtering and graph spectral properties.
In addition, empirical results indicate that Butterworth filtering serves as a more suitable low-pass filter for adaptive frequency-domain filtering compared to a direct truncation.
Naturally, when the input item embeddings are derived solely from text, 
they tend to lack sufficient collaborative information in the first.
To address this, FreLLM4Rec suggests item embeddings that incorporate both semantic features (from textual content) and collaborative features (from a pretrained model),
followed by the application of a Global Graph Low-Pass Filter to preliminarily purify the embedded collaborative signals.


Our work makes the following contributions.

\noindent$\bullet$ \textbf{Discovery of Intra-Layer Spectral Attenuation}: We identify and formally characterize this phenomenon in LLM-based recommender systems, providing theoretical explanation for observed performance gaps.

\noindent$\bullet$ \textbf{Frequency-Aware Approach}: We propose FreLLM4Rec with two synergistic modules that preserve collaborative signal integrity while addressing computational and architectural challenges.

\noindent$\bullet$ \textbf{Theoretical Foundation}: We provide rigorous analysis proving the connection between temporal frequency modulation and graph spectral property preservation.

\noindent$\bullet$ \textbf{Empirical Validation}: Extensive experiments on four datasets demonstrate improvements of up to 8.00\% in NDCG@10, validating the preservation of collaborative signals.

\section{Preliminary}
\label{sec:preliminary}
\subsection{Problem Formulation}
Let $\mathcal{U}$ denote the set of users and $\mathcal{V}$ the set of items, with cardinalities $|\mathcal{U}| = M$ and $|\mathcal{V}| = N$. Each user $u \in \mathcal{U}$ has a chronologically ordered interaction sequence $S_u = (v_1^{(u)}, v_2^{(u)}, \dots, v_{|S_u|}^{(u)})$, where $v_t^{(u)} \in \mathcal{V}$ represents the item interacted with at time $t$. The sequential recommendation task aims to predict the next item $v_{|S_u|+1}^{(u)}$ that maximizes the user's utility, formalized as learning a scoring function $f: \mathcal{U} \times \mathcal{V} \rightarrow \mathbb{R}$ that ranks all candidate items for a given user based on the interaction history.

\subsection{Graph Signal Processing}
\label{subsec:gsp_foundation}

GSP extends signal processing to irregular data~\cite{shuman2013emerging, ortega2018graph}, decomposing collaborative information into frequency components that reveal how it evolves through neural architectures.

\noindent\textbf{Item-item co-occurrence graph.} We construct an item-item graph $\mathcal{G} = (\mathcal{V}, \mathbf{W})$, where nodes represent items and edges quantify collaborative information. The adjacency matrix $\mathbf{W} \in \mathbb{R}^{N \times N}$ derives from co-occurrence: $\mathbf{W} = \mathbf{R}^T \mathbf{R}$, where $\mathbf{R} \in \mathbb{R}^{M \times N}$ is user-item interaction matrix and $W_{ij}$ counts users interacting with items $i$ and $j$. This captures the collaborative filtering principle: items that frequently appear in the same users' interaction histories exhibit similar patterns and are likely to be relevant to similar users~\cite{wang2019neural}. A graph signal $\mathbf{f} \in \mathbb{R}^N$ for a given node is taken from $\mathbf{W}$.

\noindent\textbf{Symmetric normalized Laplacian.} The graph's spectral properties are characterized by: $\mathbf{L} = \mathbf{I} - \mathbf{D}^{-1/2} \mathbf{W} \mathbf{D}^{-1/2}$, where $\mathbf{D}$ is the diagonal degree matrix with $D_{ii} = \sum_j W_{ij}$. Eigendecomposition reveals the frequency basis: $\mathbf{L} = \mathbf{U} \mathbf{\Lambda} \mathbf{U}^T$, where $\mathbf{U} = [\mathbf{u}_1, \mathbf{u}_2, \dots, \mathbf{u}_N]$ contains orthonormal eigenvectors and $\mathbf{\Lambda} = \text{diag}(\lambda_1, \lambda_2, \dots, \lambda_N)$ with $0 \leq \lambda_1 \leq \lambda_2 \leq \dots \leq \lambda_N \leq 2$.

\noindent\textbf{Graph frequencies.} Eigenvalues $\lambda_k$ represent graph frequencies. Low frequencies ($\lambda_k$ near 0) correspond to smooth signals capturing community-level patterns essential for collaborative filtering. High frequencies ($\lambda_k$ near 2) represent rapidly varying signals, often noise that hurts generalization~\cite{nt2019revisiting}. The Graph Fourier Transform (GFT) decomposes signals: $\hat{\mathbf{f}} = \mathbf{U}^T \mathbf{f}$, where $\hat{f}_k = \mathbf{u}_k^T \mathbf{f}$ is the component at frequency $\lambda_k$. The inverse transform: $\mathbf{f} = \mathbf{U} \hat{\mathbf{f}} = \sum_{k=1}^N \hat{f}_k \mathbf{u}_k$.

\noindent\textbf{Signal smoothness.} Smoothness is quantified by the Laplacian quadratic form:
\begin{equation}
\mathbf{f}^T \mathbf{L} \mathbf{f} = \sum_{i,j} W_{ij} (f_i - f_j)^2 = \sum_{k=1}^N \lambda_k |\hat{f}_k|^2.
\end{equation}
This penalizes differences between connected nodes. In spectral domain, smooth signals concentrate on energy in low frequencies. Low-frequency components capture community structures and item similarities fundamental to collaborative filtering~\cite{he2020lightgcn, wu2021self}. When these components weaken during neural processing, the system loses the ability to leverage collaborative information.

\section{LLM-based Recommendation and Spectral Analysis}
\label{sec:discovery}

This section investigates: How do pre-trained language models, optimized for linguistic tasks, process the collaborative signals that are essential for recommendation?


\subsection{The Embedding-as-Token Paradigm for LLM-based Recommendation}
\label{subsec:llm_paradigm}
The embedding-as-token paradigm~\cite{hou2024e4srec, zhang2024idgenrec} has emerged as an effective solution, enabling efficient batch processing while preserving structural information in pre-trained embeddings. Since pure text embeddings miss collaborative patterns~\cite{he2017neural} and ID-only embeddings ignore semantics, hybrid approaches combining both outperform single-modality methods~\cite{MoRec}.

To address this challenge, our approach constructs item representations that integrate both information sources. For each item $i \in \mathcal{V}$, we create a fused representation based on the following embeddings:
\begin{itemize}[leftmargin=*, noitemsep=3pt, topsep=6pt]
    \item A collaborative ID embedding $\mathbf{e}_{id}(i) \in \mathbb{R}^{d_{id}}$ pre-trained using established sequential models like SASRec~\cite{kang2018self}, encoding item popularity and co-occurrence patterns.
    \item A semantic text embedding $\mathbf{e}_{text}(i) \in \mathbb{R}^{d_{text}}$ derived from the LLM's representation of item metadata, capturing linguistic and contextual information.
\end{itemize}

These components are integrated through a learnable fusion mechanism as $\mathbf{x}_i = \text{MLP}([\mathbf{e}_{id}(i) ; \mathbf{e}_{text}(i)])$,
where MLP is a two-layer network with non-linear activation that projects the concatenated embeddings to the LLM's hidden dimension $d_{llm}$.

For a user sequence $S_u$, the LLM processes the corresponding embedding sequence:
\begin{equation}
 (\mathbf{h}_1^{(u)}, \dots, \mathbf{h}_{|S_u|}^{(u)}) = \text{LLM}(\mathbf{x}_{v_1^{(u)}}, \dots, \mathbf{x}_{v_{|S_u|}^{(u)}}).
\end{equation}

The final hidden state $\mathbf{h}_{|S_u|}^{(u)}$ serves as the user representation $\mathbf{h}_u$, and the relevance score is computed via inner product: $s(u, j) = \mathbf{h}_u^T \mathbf{x}_j$. We freeze the LLM parameters and only fine-tune the fusion MLP, allowing us to isolate and study how the pre-trained architecture affects collaborative signals without confounding factors from language model adaptation.

\subsection{Unveiling Intra-Layer Spectral Attenuation}
\label{subsec:discovery_of_decay}

Armed with the analytical approach, we now address the central question: How does the collaborative signal evolve as it propagates through LLM's deep architecture? Traditional evaluation focuses solely on final outputs, providing no insight into the internal transformations that produce them. We develop a layer-wise spectral analysis methodology to investigate signal evolution at every stage.

Our approach recognizes that Transformer self-attention creates dynamic, context-specific relationships. Therefore, rather than analyzing a single global graph, we construct local graphs tailored to each user sequence, ensuring our analysis captures the actual collaborative structure relevant to each prediction.

\begin{definition}[local Spectral Analysis approach]
For each user sequence $S_u = (v_1^{(u)}, \dots, v_{T-1}^{(u)})$, we define:
\begin{itemize}[leftmargin=*, itemsep=0pt]
    \item Local Graph Construction: We consider the subsequence $(v_2^{(u)}, v_3^{(u)}, $ $\dots, v_{T}^{(u)})$ as an ordered list of target items the model aims to predict. The adjacency matrix $\mathbf{A}_u \in \mathbb{R}^{T \times T}$ is extracted from the global co-occurrence matrix $\mathbf{W}$, where entry $A_u^{(ij)}$ is the co-occurrence weight between items $v_i$ and $v_j$ in sequence $S_u$. Repeated items are treated as distinct positional instances that share the same node identity for weight lookup. Self-loops are set to zero ($A_{ii} = 0$), and the resulting matrix is symmetrically normalized following the standard Laplacian construction described in Section~\ref{subsec:gsp_foundation}.
    
    \item Layer-wise Signal Extraction: At each layer $l$, we extract hidden states $\mathbf{H}_u^{(l)} \in \mathbb{R}^{(T-1)\times d_{llm}}$ where row $t$ contains the representation used to predict item $v_{t+1}^{(u)}$.
    
    \item Spectral Energy Analysis: The local GFT $\hat{\mathbf{H}}_u^{(l)} = \mathbf{U}_u^T \mathbf{H}_u^{(l)} (\mathbf{U}_u \in \mathbb{R}^{(T-1)\times(T-1)})$ decomposes signals into frequency components. The energy at frequency $k$ is as follows:
    \begin{equation}
    \mathcal{E}_u(l, k) = \left\|(\hat{\mathbf{H}}_u^{(l)})_k\right\|_F^2.
    \end{equation}
\end{itemize}
\end{definition}

To quantify the evolution of collaborative signals within LLM architectures, we aggregate energy measurements across users and partition frequencies into quantile-based bands. Through this systematic spectral characterization, we identify a consistent phenomenon illustrated in Figure \ref{fig:decay_phenomenon_detailed}: The energy of low-frequency collaborative signals monotonically decreases as representations propagate to deeper layers. This systematic attenuation of the smoothest, most essential collaborative patterns represents a failure mode of LLM architectures for recommendation.

As a comparison, we also conduct identical analysis on SASRec~\cite{kang2018self}, a Transformer specifically designed for sequential recommendation. As shown in Figure~\ref{fig:sasrec_preserved}, SASRec preserves low-frequency energy throughout its layers, demonstrating that spectral attenuation is not inherent to the Transformer architecture but a consequence of the inductive biases from language-oriented pre-training. The LLM's self-attention mechanism, while adept at semantics, systematically erodes the vital, smooth collaborative patterns represented by low-frequency graph signals.

This discovery has important implications. The low-frequency components that LLMs systematically attenuate are precisely those encoding community preferences, item similarity patterns, and collaborative filtering signals—the foundation of effective recommendation. This phenomenon explains why many LLM-based recommendation systems fail to fully leverage collaborative information despite their powerful semantic understanding capabilities. It motivates the development of our frequency-aware correction mechanisms presented in the next section.

\begin{figure}[!t]
  \centering
  \includegraphics[width=1.0\columnwidth]{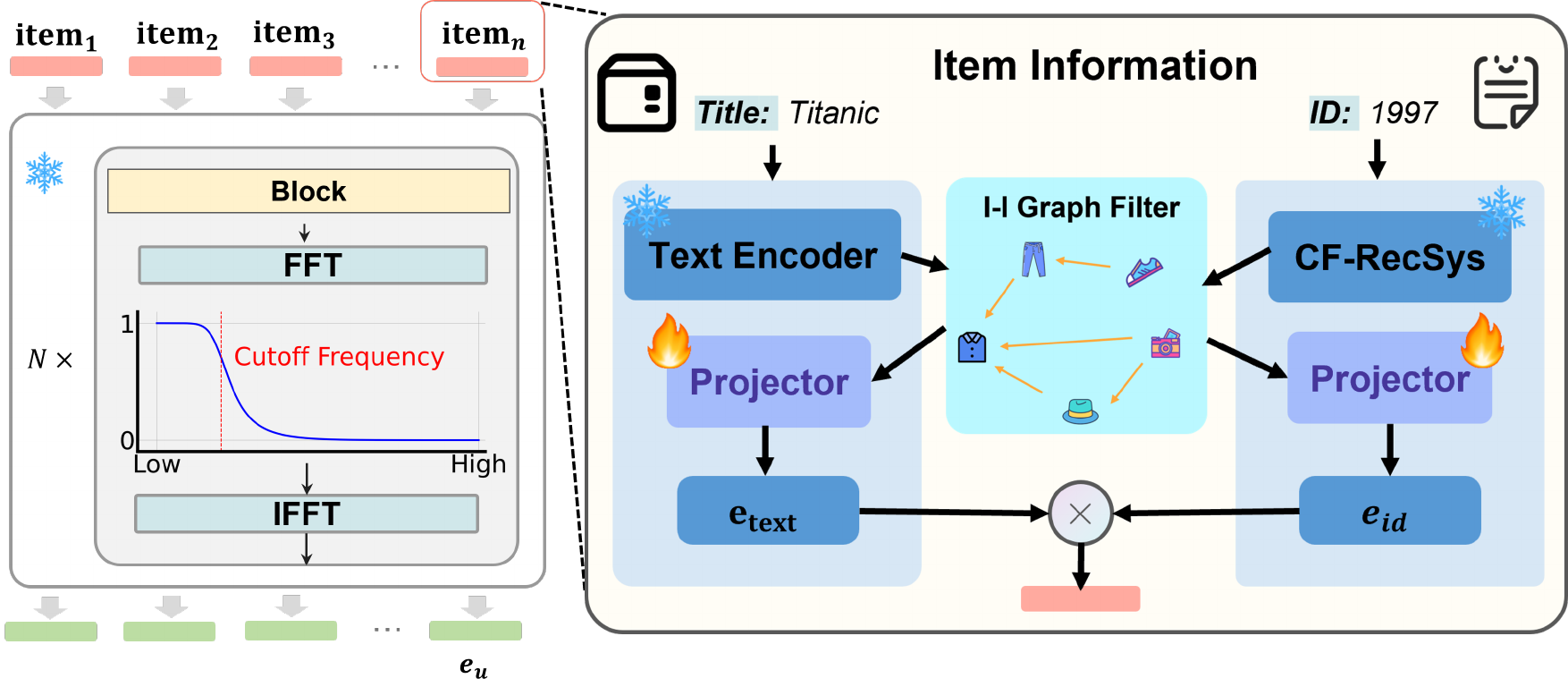}
  \caption{Overview of FreLLM4Rec: Our frequency-aware approach transforms spectral decay into spectral preservation, with the G-LPF module purifying input collaborative signals and the TFM module actively counteracting attenuation after each Transformer layer.}
  \label{fig:framework}
\end{figure}

\section{The Proposed FreLLM4Rec Approach}
\label{sec:methodology}

Motivated by the discovery of the spectral attenuation phenomenon, we present FreLLM4Rec (Frequency-aware LLM for Recommendation), designed to preserve collaborative signal integrity throughout the LLM processing pipeline. Our approach introduces two synergistic frequency-aware modules: input-level signal purification and intra-layer signal preservation, which operate at complementary stages within the LLM-based recommendation framework. Figure \ref{fig:framework} illustrates the complete architecture.

\subsection{Input Signal Purification via Global Graph Low-Pass Filtering}
\label{subsec:g_lpf}

\begin{figure}[!t]
    \centering
    \includegraphics[width=\linewidth]{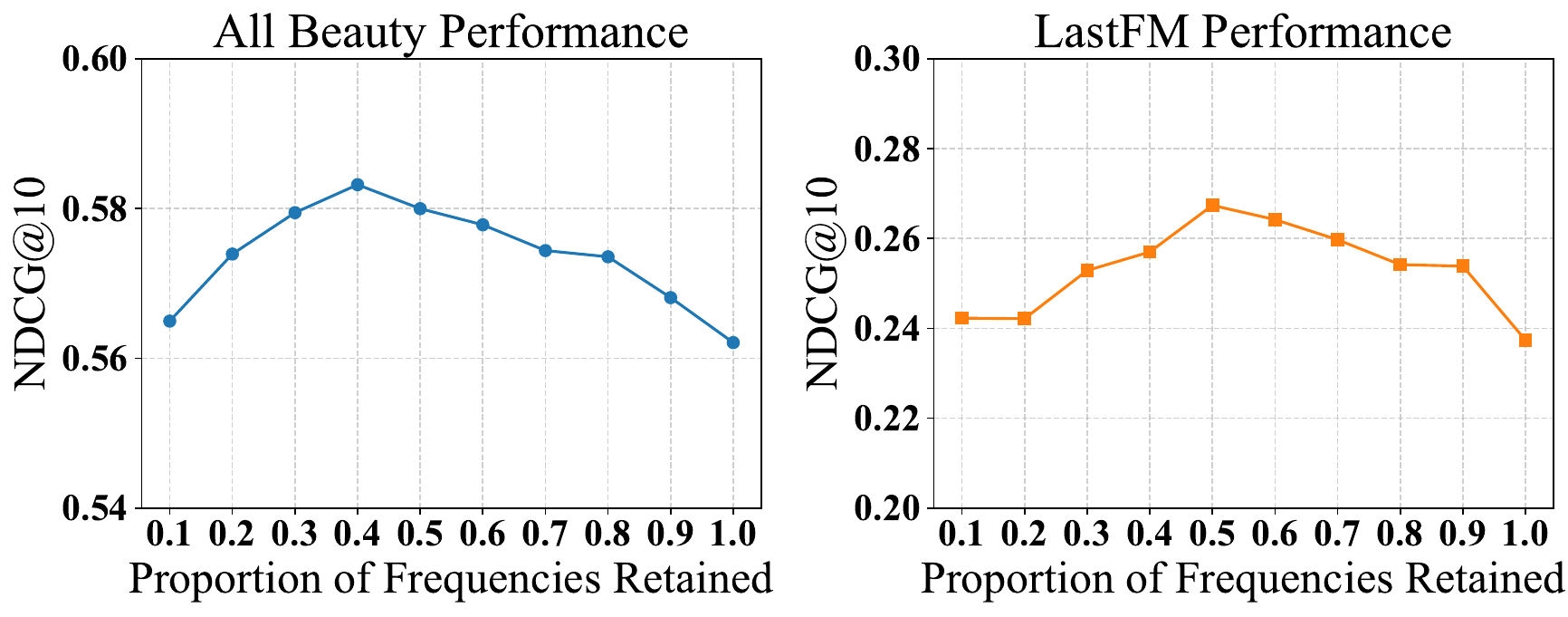}
    \caption{Impact of retaining percentages of low-frequency components on NDCG@10.}
    \label{fig:performance_comparison_larger_font}
\end{figure}

While pre-trained collaborative embeddings encode valuable interaction patterns, they may also contain less stable high-frequency components that could exacerbate the attenuation problem within LLMs. Our G-LPF operates on the global item-item co-occurrence graph $\mathcal{G}$ (defined in Section ~\ref{subsec:gsp_foundation}) that captures collaborative relationships across the entire item catalog. To investigate the impact of frequency composition on recommendation performance, we applied an ideal low-pass filter, progressively retaining larger portions of the low-frequency components of the ID embeddings. The results, shown in Figure~\ref{fig:performance_comparison_larger_font}, revealed a consistent pattern: performance initially improved as more low-frequency components were included, but peaked and subsequently degraded upon the inclusion of the highest-frequency components. This provides empirical evidence that the highest-frequency components contribute less to recommendation effectiveness, motivating the need for an input-level purification module.

This observation motivates our first intervention: a G-LPF that purifies collaborative signals before they enter LLMs. Low frequency signals represent smooth variations where similar items have similar representations—the essence of collaborative filtering~\cite{he2020lightgcn}. High frequency components often encode popularity biases, measurement noise, or spurious correlations that hinder generalization~\cite{Min2020ScatteringGO}. By selectively attenuating these potentially detrimental high-frequency components, we provide LLMs with cleaner collaborative signals.

The ideal spectral filtering operation would be:
\begin{equation}
 \mathbf{E}'_{item} = \mathbf{U} \, \text{diag}\big(h(\lambda_1), \dots, h(\lambda_N)\big) \, \mathbf{U}^T \mathbf{E}_{item},
 \label{eq:ideal_gft_filter}
\end{equation}
where $h(\lambda)$ is a low-pass frequency response function. However, this requires $\mathcal{O}(N^3)$ eigendecomposition, which is computationally prohibitive for large item catalogs in recommender systems.

We leverage a result from spectral graph theory: any spectral filter can be approximated by a polynomial of the Laplacian ~\cite{defferrard2016convolutional}:
\begin{equation}
 h(\lambda) = \sum_{k=0}^{K} \theta_k \lambda^k \implies \mathcal{H}(\mathbf{L}) = \sum_{k=0}^{K} \theta_k \mathbf{L}^k.
 \label{eq:poly_response}
\end{equation}

Our G-LPF applies this operator to purify ID embeddings:
\begin{equation}
 \mathbf{E}'_{item} = \mathcal{H}(\mathbf{L}) \mathbf{E}_{item}.
 \label{eq:g_lpf_final_general}
\end{equation}

The sparse structure of collaborative graphs in real-world ensures computational efficiency. In our experiments, we adopt a simplified first-order filter $(K=1)$ with response $h(\lambda) = 1 - \alpha\lambda$, where $\alpha \in [0,1]$ controls filtering strength, with $\alpha = 0$ corresponding to no filtering and $\alpha = 1$ providing maximum smoothing.

\subsection{Intra-Layer Temporal Frequency Modulation for Signal Preservation}
\label{subsec:tfm}

While input purification helps, it cannot prevent systematic attenuation within LLM layers. To address this, we employ Temporal Frequency Modulation (TFM), which operates after each Transformer layer to restore degraded low-frequency components. However, developing effective intra-layer correction presents challenges. Firstly, direct local graph-domain operations require expensive eigendecomposition at each layer, making them computationally intractable. 
Secondly, naive spectral filtering cannot leverage temporal dependencies in user sequences. This is further compounded by maintaining compatibility with pre-trained LLMs while ensuring corrections don't interfere with semantic processing.

The key insight enabling TFM is that temporal and graph frequency domains can be connected under certain conditions: The Graph Fourier Transform on a ring graph is equivalent to the Discrete Fourier Transform (see Appendix~\ref{subsec:dft_gft_relationship}). While user sequences do not form perfect ring graphs, temporal filtering can influence graph spectral properties through:

\begin{assumption}[Spatio-Temporal Locality]
\label{assump:locality}
Items appearing in temporal proximity within user sequences exhibit higher collaborative similarity (larger edge weights) than temporally distant items.
\end{assumption}

This assumption, validated across recommendation domains~\cite{he2016fusing,rendle2010factorizing,mcauley2013hidden}, reflects smooth preference evolution, i.e., users interact with similar items in succession. Based on this, we can establish the connection between temporal and graph frequency domains:

\begin{theorem}[Informal]
\label{thm:temporal_to_graph}
Under Assumption \ref{assump:locality}, temporal low-pass filtering of item representation sequences reduces their graph Laplacian quadratic form and concentrates energy in low graph frequencies.
\end{theorem}

Theorem~\ref{thm:temporal_to_graph} (proved in Appendix \ref{app:proof_theorem}) provides TFM's foundation: smoothing hidden states temporally restores low-frequency graph components while avoiding computational overhead of direct graph operations.

To preserve collaborative signals while avoiding ringing artifacts that could distort sequential patterns, we implement TFM using the Butterworth filter. The Butterworth design is particularly suitable for our application due to its maximally flat passband response, which preserves low-frequency collaborative components without distortion, and its smooth roll-off characteristic that gradually attenuates high frequencies without introducing phase discontinuities that could corrupt sequential dependencies~\cite{Champagne2004DiscreteTS}. Compared to hard frequency cutoffs that create ringing artifacts or Gaussian filters that may over-smooth temporal dynamics, Butterworth filtering achieves an optimal balance between collaborative signal preservation and sequential pattern integrity:
\begin{equation}
 \mathbf{H}'^{(l)} = \text{TFM}(\mathbf{H}^{(l)}) = \mathcal{F}^{-1} \left( \mathcal{B}(\omega) \odot \mathcal{F}(\mathbf{H}^{(l)}) \right),
 \label{eq:tfm_operator}
\end{equation}
where $\mathcal{F}$ denotes 1D FFT along the sequence dimension, and Butterworth filter is as follows:
\begin{equation}
 |\mathcal{B}(\omega)|^2 = \frac{1}{1 + (\omega / \omega_c)^{2n}}.
 \label{eq:butterworth_filter}
\end{equation}
Here $\omega_c$ and $n$ are cutoff frequency and order hyperparameters.

TFM applies after each Transformer layer, creating continuous correction that counteracts progressive attenuation. This ensures collaborative signals remain strong throughout network depth, enabling LLMs to leverage both semantic understanding and preserved collaborative patterns.

\begin{table*}[!t]
\centering
\caption{Overall performance comparison, with the best results in bold and best baseline underlined. 'Improv.' denotes the relative improvement of FreLLM4Rec over the best baseline.}
\label{tab:main_results}
\begin{tabular}{@{}c|cc|cc|cc|cc@{}}
\toprule
\multirow{2}{*}{Model} & \multicolumn{2}{c|}{All Beauty} & \multicolumn{2}{c|}{Luxury Beauty} & \multicolumn{2}{c|}{Movies and TV} & \multicolumn{2}{c}{LastFM} \\
& NDCG@10 & Recall@10 & NDCG@10 & Recall@10 & NDCG@10 & Recall@10 & NDCG@10 & Recall@10 \\
\midrule
GRU4Rec & 0.5379 & 0.5904 & 0.4352 & 0.5797 & 0.5350 & 0.7531 & 0.1969 & 0.3440 \\
Caser & 0.5322 & 0.5895 & 0.3739 & 0.5104 & 0.5073 & 0.7013 & 0.1042 & 0.2073 \\
SASRec & 0.5774 & 0.6127 & 0.5069 & 0.6275 & 0.3566 & 0.5118 & 0.1996 & 0.3502 \\
Bert4Rec & 0.5370 & 0.5811 & 0.3961 & 0.4951 & 0.5486 & 0.7674 & 0.1782 & 0.3189 \\
MoRec & 0.5551 & 0.5809 & 0.4780 & 0.6209 & 0.4015 & 0.6260 & 0.2000 & 0.3710 \\
\midrule
\midrule
FMLPRec & 0.5662 & 0.6110 & 0.5065 & 0.6385 & 0.4440 & 0.6384 & 0.2185 & 0.3800 \\
BSARec & 0.5634 & 0.6013 & 0.4902 & 0.6081 & 0.3722 & 0.5760 & 0.2214 & 0.3862 \\
\midrule
\midrule
SR-GNN & 0.5502 & 0.5986 & 0.4274 & 0.5604 & 0.3909 & 0.5648 & 0.1563 & 0.2902 \\
MAERec & 0.5772 & 0.6145 & 0.4869 & 0.5982 & 0.4109 & 0.5996 & 0.1702 & 0.2911 \\
\midrule
\midrule
LLaMA-3 & 0.1372 & 0.2527 & 0.0824 & 0.2033 & 0.1643 & 0.2824 & 0.0872 & 0.2147 \\
LLARA & 0.5418 & 0.5911 & 0.5072 & 0.6324 & 0.5346 & 0.7606 & 0.2810 & 0.4955 \\
E4SRec & 0.5415 & 0.5850 & 0.5120 & 0.6330 & 0.5550 & 0.7610 & 0.2305 & 0.3901 \\
IDGenRec & \underline{0.5821} & \underline{0.6198} & 0.5299 & 0.6412 & 0.3950 & 0.6100 & 0.2655 & 0.4705 \\
LLM2Rec & 0.5624 & 0.6071 & \underline{0.5354} & \underline{0.6590} & \underline{0.5976} & \underline{0.7812} & \underline{0.3097} & \underline{0.5119} \\
\midrule
\midrule
FreLLM4Rec (Ours) & \textbf{0.6287} & \textbf{0.6892} & \textbf{0.5618} & \textbf{0.7020} & \textbf{0.6311} & \textbf{0.8163} & \textbf{0.3327} & \textbf{0.5462} \\
\textit{Improv.} & 8.00\% & 11.20\% & 4.93\% & 6.52\% & 5.61\% & 4.49\% & 7.43\% & 6.70\% \\
\bottomrule
\end{tabular}%
\end{table*}

\subsection{Computational Efficiency and Scalability}
\label{subsec:complexity}
FreLLM4Rec is designed for practical deployment at scale. The G-LPF preprocessing has complexity $\mathcal{O}(K \cdot |\mathcal{E}_{\text{item}}| \cdot d)$, where $K$ is the polynomial order, $|\mathcal{E}_{\text{item}}|$ is the number of edges in the sparse item graph, and $d$ is the embedding dimension. This preprocessing represents a one-time cost that is amortized across the entire training process.

The TFM module adds $\mathcal{O}(B \cdot d \cdot T \log T)$ computational cost per layer, where $B$ is the batch size, $d$ is the hidden dimension, and $T$ is the sequence length. This complexity is asymptotically smaller than the Transformer's $\mathcal{O}(B \cdot T^2 \cdot d)$ self-attention complexity and the feed-forward network's $\mathcal{O}(B \cdot T \cdot d^2)$ complexity. Empirical timing analysis (detailed in Appendix~\ref{app:timing_analysis}) confirms that TFM introduces minimal overhead (approximately 3\% increase in training time) while achieving substantial performance improvements. Consequently, FreLLM4Rec maintains the same scalability characteristics as standard LLM architectures while providing frequency-aware corrections.

\section{Experiments}
\label{sec:experiments}

\subsection{Experimental Settings}

\textbf{Datasets.} We evaluate on four benchmark datasets spanning e-commerce and entertainment domains: All Beauty, Luxury Beauty, and Movies and TV from the Amazon product data\footnote{\url{https://cseweb.ucsd.edu/~jmcauley/datasets/amazon_v2/}} ~\cite{ni2019justifying}, and LastFM\footnote{\url{https://grouplens.org/datasets/hetrec-2011/}} for music recommendation ~\cite{cantador2011second}. Following standard protocols ~\cite{kang2018self, zhou2020s3}, we filter users and items with fewer than 5 interactions and use the leave-one-out strategy for train/valid/test splits.


\noindent\textbf{Baselines.} To validate the effectiveness of our approach, we compare it with several research lines: 
I) Traditional sequential models: GRU4Rec~\cite{hidasi2015session}, Caser~\cite{tang2018personalizedtopnsequentialrecommendation}, SASRec~\cite{kang2018self}, BERT4Rec~\cite{sun2019bert4rec}, MoRec~\cite{MoRec}; 
II) Frequency-domain sequential models: FMLPRec~\cite{zhou2022filter}, BSARec~\cite{shin2024attentiveinductivebiassequential}; 
III) Graph-based sequential models: SR-GNN~\cite{wu2019session}, MAERec~\cite{Ye_2023}; 
IV) LLM-based recommendation methods: LLaMA-3~\cite{touvron2023llama}, LLARA~\cite{liao2024llara}, E4SRec~\cite{hou2024e4srec}, IDGenRec~\cite{zhang2024idgenrec}, LLM2Rec~\cite{he2025llm2reclargelanguagemodels}. 

\noindent\textbf{Implementation Details.} All experiments use PyTorch on NVIDIA A800 GPUs. For FreLLM4Rec and LLM baselines, we use Qwen2.5-7B-Instruct ~\cite{yang2024qwen2} as the default backbone. Throughout all experiments, we freeze the LLM backbone parameters and only train the lightweight fusion MLP, allowing us to isolate and study how pre-trained architectures process collaborative signals. We employ AdamW optimization with batch size 32, searching learning rates in $\{1e-5, 5e-5, 1e-4, 5e-4\}$. The G-LPF parameter $\alpha$ is tuned in $[0, 1.0]$ and TFM cutoff $\omega_c$ in $[0, 1.0]$. For fair comparison, all models use 50-dimensional ID embeddings. For LLM-based methods that leverage pre-trained collaborative ID embeddings (e.g., E4SRec, LLARA), these 50-dimensional embeddings are projected into the LLM’s hidden space via a trainable fusion MLP.

\begin{table}[tbp]
\footnotesize
\centering
\caption{Ablation study of component contributions.}
\label{tab:ablation_study}
\begin{tabular}{@{}c|cc|cc|cc@{}}
\toprule
\multirow{2}{*}{Model Variant} & \multicolumn{2}{c|}{LastFM} & \multicolumn{2}{c|}{All Beauty} & \multicolumn{2}{c}{Luxury Beauty} \\
\cmidrule(lr){2-3} \cmidrule(lr){4-5} \cmidrule(lr){6-7}
& N@10 & R@10 & N@10 & R@10 & N@10 & R@10 \\
\midrule
FreLLM4Rec (Full) & \textbf{0.3327} & \textbf{0.5462} & \textbf{0.6287} & \textbf{0.6892} & \textbf{0.5618} & \textbf{0.7020} \\
\midrule
\midrule
w/o G-LPF & 0.2524 & 0.4523 & 0.5958 & 0.6671 & 0.5336 & 0.6575 \\
w/o TFM & 0.2507 & 0.4468 & 0.5840 & 0.6392 & 0.5181 & 0.6408 \\
w/o G-LPF \& TFM & 0.2236 & 0.4037 & 0.5564 & 0.6090 & 0.5059 & 0.6257 \\
\midrule
\midrule
w/o ID emb & 0.0629 & 0.1404 & 0.5378 & 0.6134 & 0.4954 & 0.5946 \\
w/o Text emb & 0.2374 & 0.4257 & 0.5355 & 0.5909 & 0.4478 & 0.5967 \\
\bottomrule
\end{tabular}
\end{table}

\noindent\textbf{Evaluation Metrics.} We report Recall@10 and NDCG@10 as our primary evaluation metrics. For each sequence, we randomly select 100 non-interacted items to construct the candidate set, ensuring the inclusion of the correct subsequent item. This 100-item candidate setting is adopted to enable fair comparison with existing large language model-based recommendation methods. For traditional models, we select the candidate item with the highest probability as the prediction, while LLM-based models generate predictions through their respective inference mechanisms.

\subsection{Overall Performance Comparison}

Table \ref{tab:main_results} presents the performance comparison. Our method achieves competitive results across all datasets and metrics, with notable improvements on LastFM (7.43\% NDCG@10 improvement) and All Beauty(8.00\% improvement). Several key observations emerge:

First, FreLLM4Rec outperforms all LLM-based baselines, demonstrating that addressing spectral attenuation is important for effective LLM-based recommendation. The performance difference is substantial—while methods like E4SRec and IDGenRec show modest improvements over traditional models, FreLLM4Rec achieves larger performance gains.

Second, our approach surpasses specialized frequency-domain models like FMLPRec and BSARec, validating that our targeted correction of LLM-specific spectral attenuation is more effective than generic frequency-aware designs.

Third, the consistent improvements across diverse domains (beauty products, movies, music) demonstrate the generality of both the spectral attenuation phenomenon and our solution.

\subsection{Component Analysis and Spectral Decay Mitigation}
\label{subsec:ablation}


\noindent\textbf{Component Contributions.} Table \ref{tab:ablation_study} demonstrates that both modules are essential: removing TFM causes 6-8\% performance drop, while removing G-LPF leads to 3-5\% degradation. The full model's performance exceeds the sum of individual contributions, indicating module synergy. Removing ID embeddings causes up to 60\% performance loss, confirming collaborative signals remain essential despite LLMs' semantic capabilities. We additionally verify that the observed attenuation is not attributable to insufficient fusion MLP capacity: as shown in Table~\ref{tab:mlp_capacity}, varying MLP depth and width produces less than 0.5\% NDCG@10 change across all configurations.

\begin{table}[!t]
\centering
\caption{NDCG@10 of different fusion MLP capacities.}
\label{tab:mlp_capacity}
\begin{tabular}{lccc}
\toprule
MLP Variant & All Beauty & Luxury Beauty & LastFM \\
\midrule
2-layer (default)       & 0.398 & 0.220 & 0.022 \\
2-layer $\times$2 width & 0.399 & 0.221 & 0.022 \\
4-layer                 & 0.398 & 0.219 & 0.023 \\
\bottomrule
\end{tabular}
\end{table}

\begin{figure}[!t]
    \centering
    \begin{subfigure}[b]{0.48\linewidth}
        \centering
        \includegraphics[width=\textwidth]{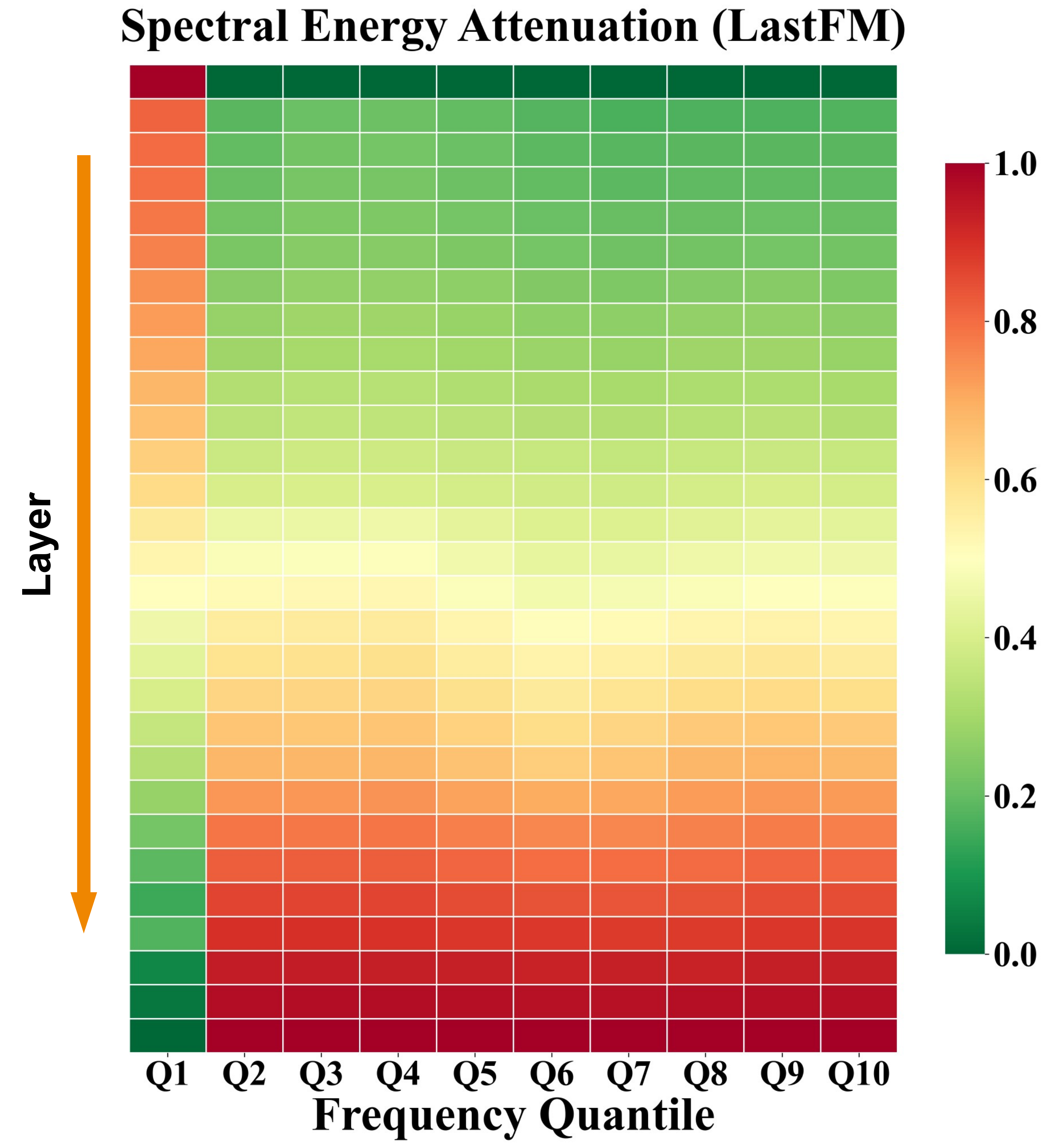}
        \caption{Vanilla LLM (LastFM)}
        \label{fig:vanilla_llm_decay_lastfm}
    \end{subfigure}
    \hfill
    \begin{subfigure}[b]{0.48\linewidth}
        \centering
        \includegraphics[width=\textwidth]{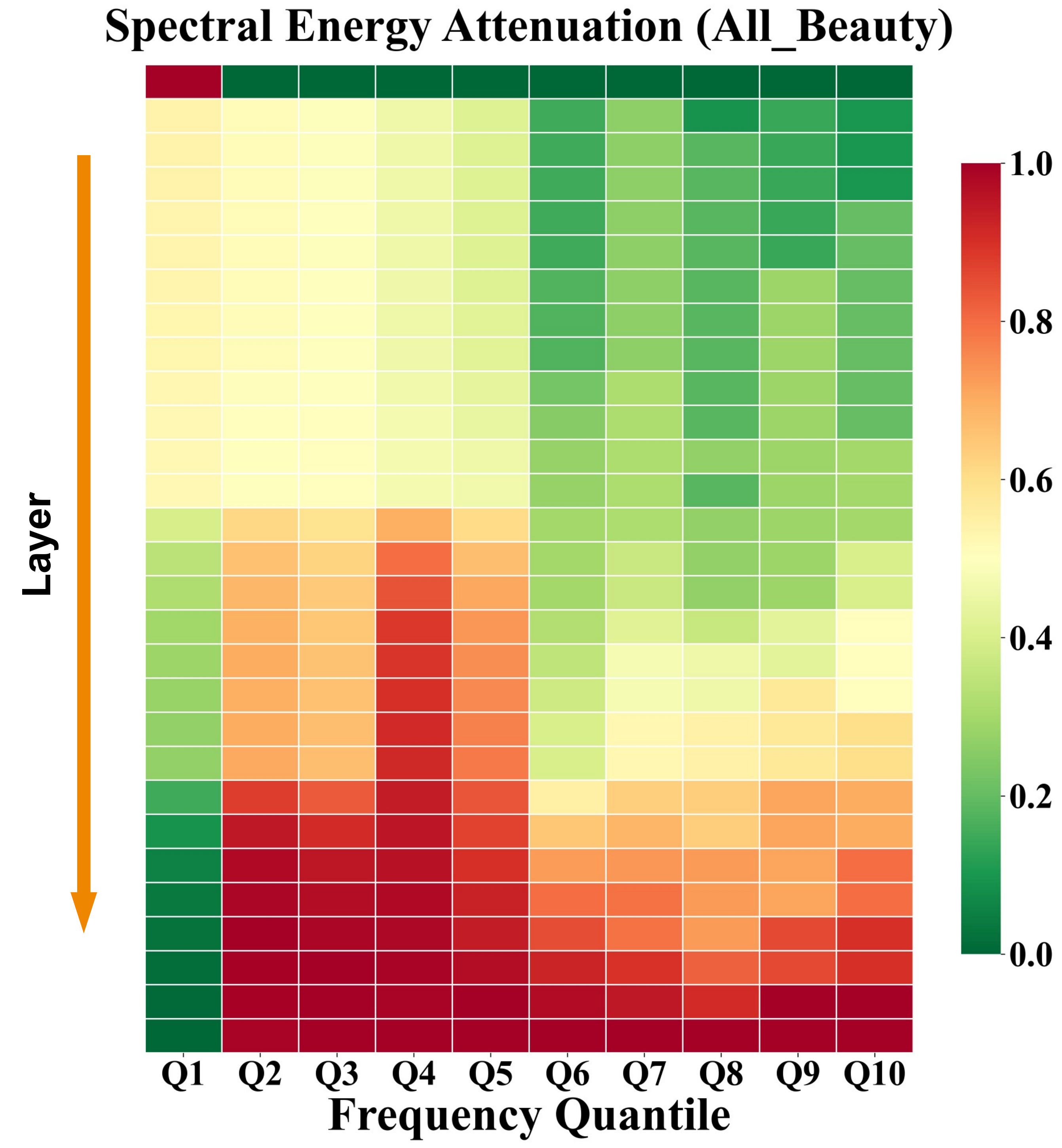}
        \caption{Vanilla LLM (All Beauty)}
        \label{fig:vanilla_llm_decay_beauty}
    \end{subfigure}
    \\
    \begin{subfigure}[b]{0.48\linewidth}
        \centering
        \includegraphics[width=\textwidth]{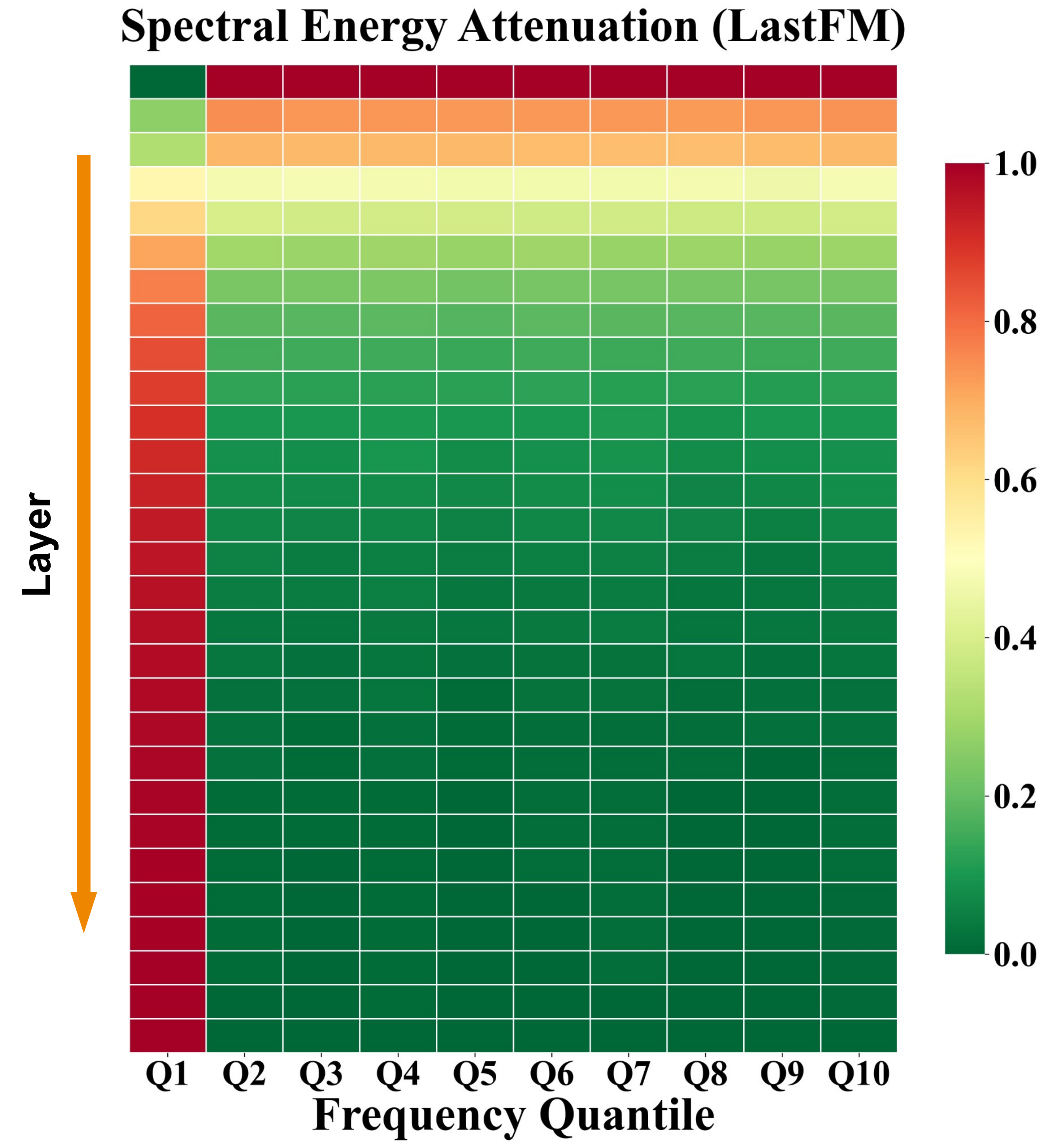}
        \caption{FreLLM4Rec (LastFM)}
        \label{fig:frellm4rec_preserved_lastfm}
    \end{subfigure}
    \hfill
    \begin{subfigure}[b]{0.48\linewidth}
        \centering
        \includegraphics[width=\textwidth]{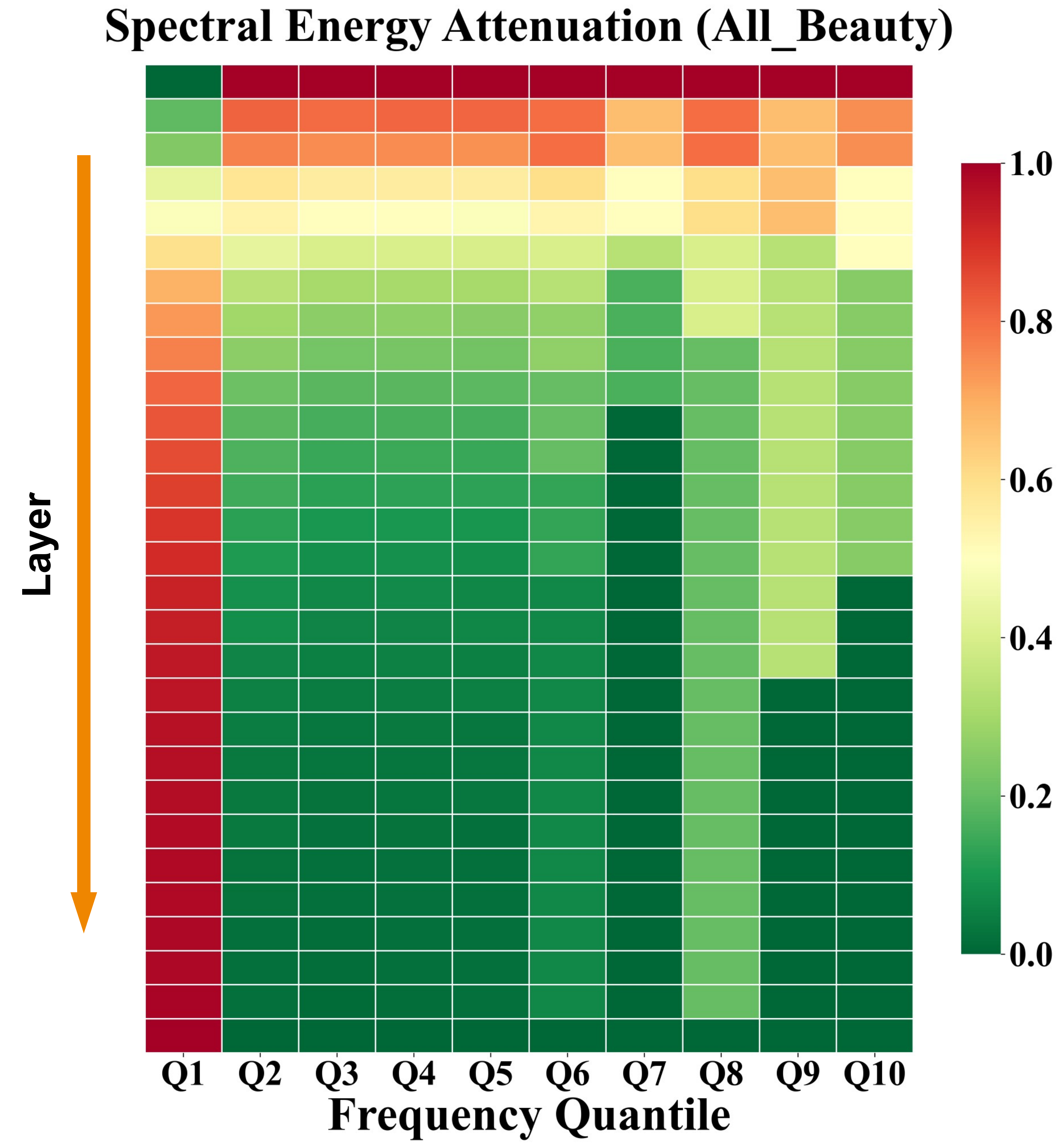}
        \caption{FreLLM4Rec (All Beauty)}
        \label{fig:frellm4rec_preserved_beauty}
    \end{subfigure}
    \caption{Validation of spectral decay mitigation. 
    Frequency-wise (column-wise) min-max normalization is applied to highlight 
    relative cross-layer trends for each frequency component, making the evolution patterns more visible.
    }
    \label{fig:spectral_mitigation}
\end{figure}

\noindent\textbf{Empirical Spectral Preservation.} Figure \ref{fig:spectral_mitigation} validates FreLLM4Rec effectively preserves low-frequency collaborative energy across layers, while vanilla LLMs exhibit severe attenuation with over 70\% energy degradation by the final layers. This preservation contributes to our method's superior recommendation performance.

\begin{table}[!t]
\centering
\caption{Performance (NDCG@10) comparison of vanilla LLM backbones (Base) with FreLLM4Rec (+Fre).}
\label{tab:backbone_ablation}
\resizebox{\columnwidth}{!}{%
\begin{tabular}{@{}c|cc|cc|cc@{}}
\toprule
\multirow{2}{*}{LLM Backbone} & \multicolumn{2}{c|}{All Beauty} & \multicolumn{2}{c|}{Movies and TV} & \multicolumn{2}{c}{LastFM} \\
& Base & +Fre & Base & +Fre & Base & +Fre \\
\midrule
Qwen2.5-7B      & 0.5564 & \textbf{0.6287} & 0.6025 & \textbf{0.6311} & 0.2236 & \textbf{0.3327}\\
Llama3.1-8B     & 0.5671 & \textbf{0.5886} & 0.6188 & \textbf{0.6402} & 0.2341 & \textbf{0.3174}\\
Mistral-7B-v0.3 & 0.5794 & \textbf{0.6117} & 0.5961 & \textbf{0.6444} & 0.1776 & \textbf{0.3225}\\
\bottomrule
\end{tabular}
}
\end{table}
\subsection{Robustness Analysis Across Architectures}

Table~\ref{tab:backbone_ablation} shows that both Llama-3.1-8B and Mistral-7B suffer similar spectral degradation patterns. FreLLM4Rec consistently achieves 4-7\% gains with stable G-LPF and TFM contribution ratios across all tested backbones, confirming that the approach addresses a fundamental architectural limitation.

Table \ref{tab:id_network_impact} shows robustness to different collaborative signal sources. Whether using embeddings from SASRec, MAERec, or BSARec, FreLLM4Rec consistently improves performance by 3-6\%. This confirms our approach addresses an architectural issue rather than artifacts of specific embedding methods.

\subsection{Hyperparameter Analysis and Sensitivity}

\begin{figure}[h]
    \centering
    \includegraphics[width=\linewidth]{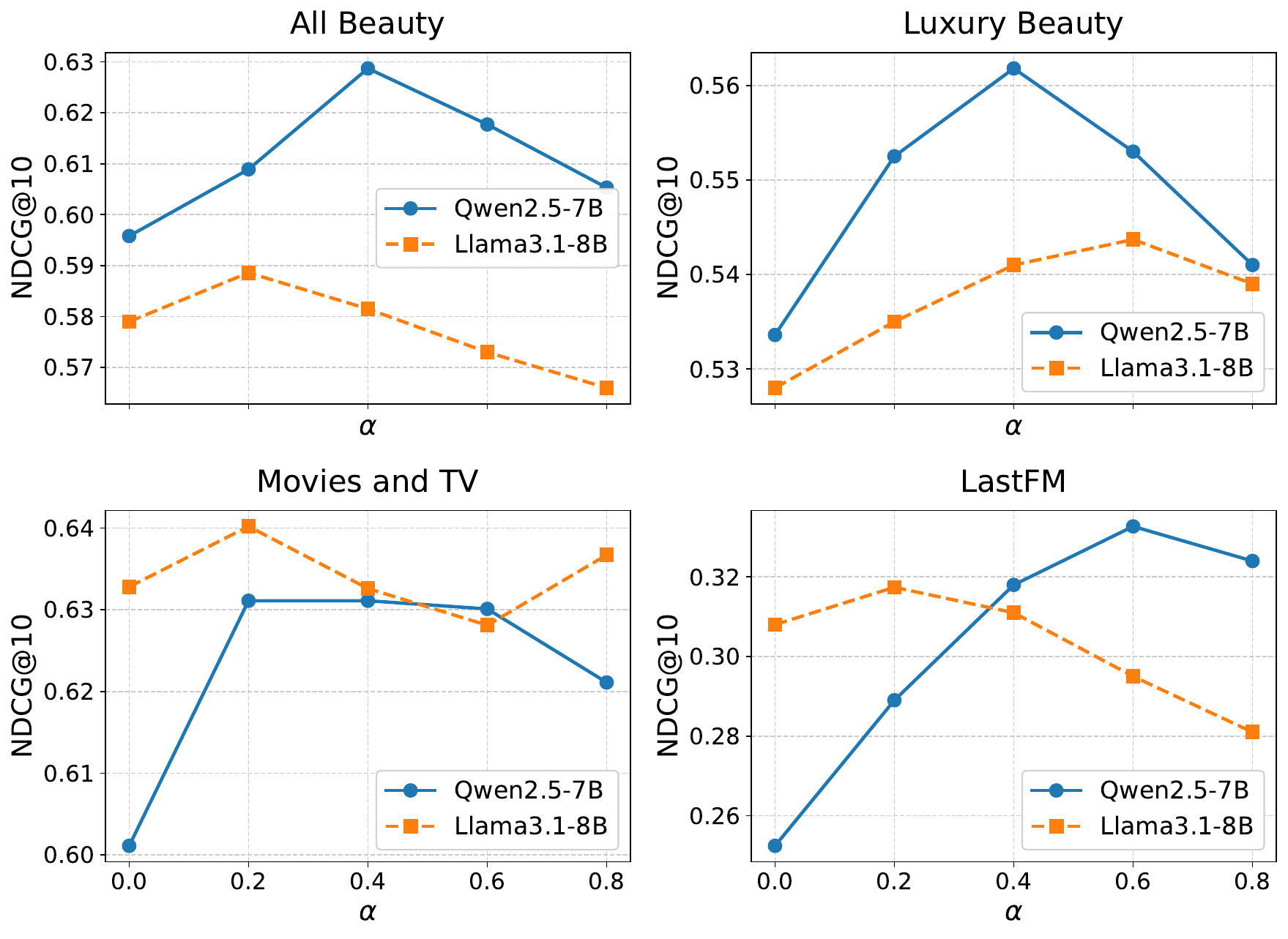}
    \caption{G-LPF filtering strength analysis. 
    }
    \label{fig:alpha_sensitivity}
\end{figure}

Figure \ref{fig:alpha_sensitivity} shows the impact of G-LPF strength $\alpha$. Performance improves as we increase filtering up to $\alpha \approx 0.3$, confirming that removing high-frequency noise helps. Beyond this point, aggressive filtering begins removing useful signals, causing performance degradation. The optimal range (0.2-0.5) indicates robustness.

\begin{figure}[h]
    \centering
    \includegraphics[width=\linewidth]{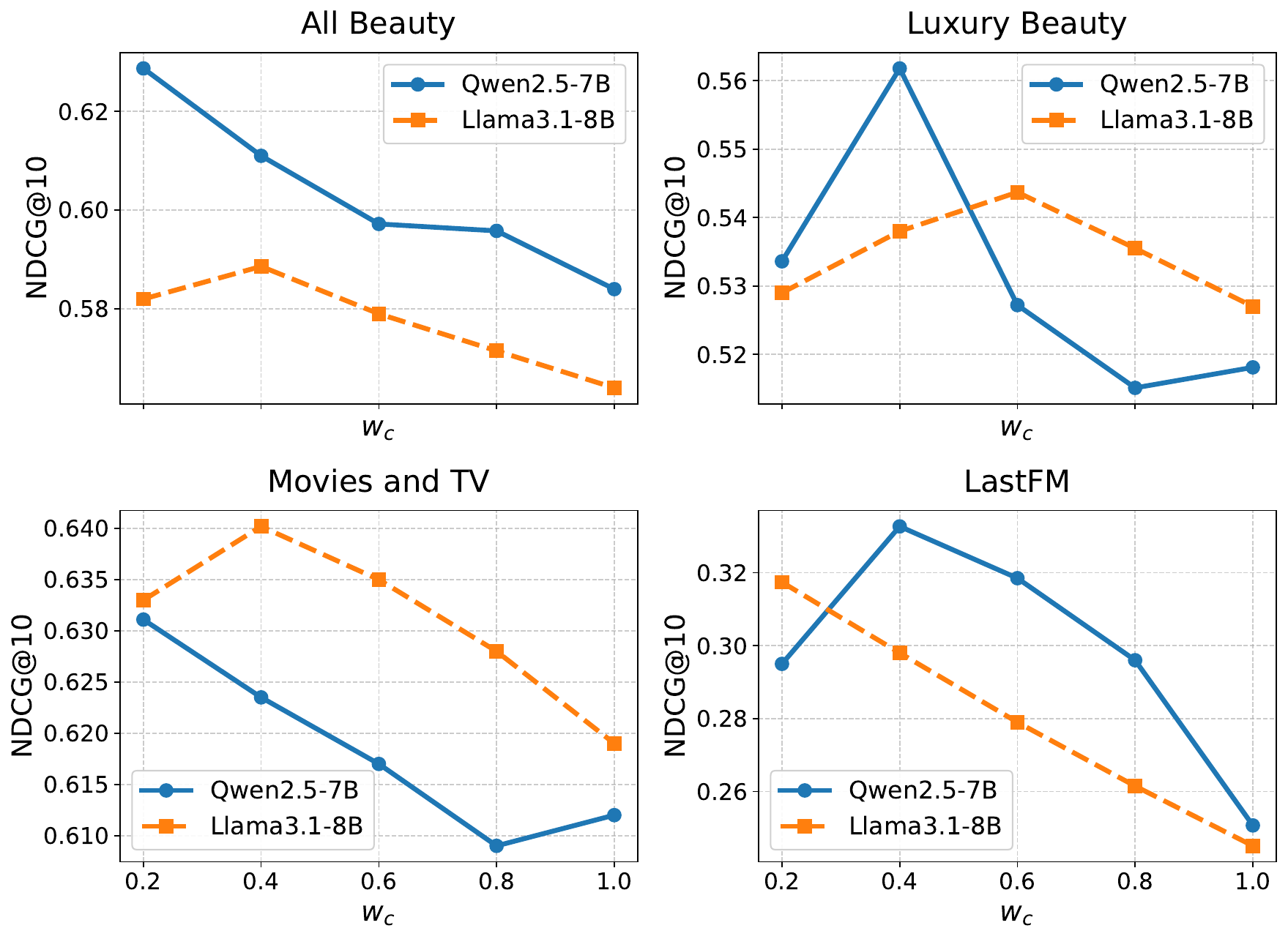}
    \caption{TFM cutoff frequency analysis. 
    }
    \label{fig:omega_sensitivity}
\end{figure}

Figure \ref{fig:omega_sensitivity} analyzes TFM cutoff frequency $\omega_c$. Without filtering ($\omega_c = 1.0$), spectral attenuation impacts performance. As we decrease $\omega_c$, performance improves by preserving low frequencies. The optimal range (0.1-0.5) balances preserving collaborative patterns with maintaining necessary sequential dynamics.

\subsection{Performance Across LLM Scales}
\label{app:llm_scales}
We evaluate FreLLM4Rec with multiple LLM backbones of varying parameter scales. We report NDCG@10 on All Beauty, Movies and TV, and LastFM.

Across both Llama and Qwen families, performance shows a clear positive correlation with model size. The gains are particularly notable on All Beauty and LastFM, indicating that the frequency-aware design scales well with larger and deeper backbones while continuing to preserve collaborative signals.

\begin{table}[htbp]
\centering
\caption{Performance (NDCG@10) comparison between traditional ID-based models (Trad.) and FreLLM4Rec (Fre.) using embeddings from different sequential recommenders.}
\label{tab:id_network_impact}
\resizebox{.95\columnwidth}{!}{%
\begin{tabular}{@{}c|cc|cc|cc@{}}
\toprule
\multirow{2}{*}{ID Network} & \multicolumn{2}{c|}{All Beauty} & \multicolumn{2}{c|}{Movies and TV} & \multicolumn{2}{c}{LastFM} \\
& Trad. & Fre. & Trad. & Fre. & Trad. & Fre. \\
\midrule
SASRec & 0.5774 & \textbf{0.6287} & 0.3566 & \textbf{0.6311} & 0.1996 & \textbf{0.3327} \\
MAERec & 0.5772 & \textbf{0.6020} & 0.4109 & \textbf{0.6012} & 0.1702 & \textbf{0.2793} \\
BSARec & 0.5634 & \textbf{0.6112} & 0.3722 & \textbf{0.5983} & 0.2214 & \textbf{0.3062} \\
\bottomrule
\end{tabular}%
}
\end{table}

\subsection{Impact of LLM Fine-tuning on Collaborative Signal Preservation}
\label{app:lora_finetuning}

To investigate whether standard fine-tuning approaches can mitigate the spectral attenuation phenomenon, we conducted additional experiments using LoRA (Low-Rank Adaptation)~\cite{Hu2021LoRALA}, a widely-adopted parameter-efficient fine-tuning method. We apply LoRA modules with rank $r=8$ and scaling factor $\alpha=16$ to all attention projection layers, training for 5 epochs.

\begin{table}[!t]
\centering
\caption{Performance (NDCG@10) comparison between vanilla LLM(Base), LoRA fine-tuned LLM, and FreLLM4Rec across different backbones.}
\label{tab:lora_performance}
\begin{tabular}{llccc}
\toprule
LLM & Variant & All Beauty & Movies & LastFM \\
\midrule
\multirow{3}{*}{Llama-3.1-8B} 
& Base & 0.5671 & 0.6188 & 0.2341 \\
& +LoRA & 0.5632 & 0.6013 & 0.2581 \\
& +FreLLM4Rec & \textbf{0.5886} & \textbf{0.6402} & \textbf{0.3174} \\
\midrule
\multirow{3}{*}{Qwen2.5-7B} 
& Baseline & 0.5564 & 0.6025 & 0.2236 \\
& +LoRA & 0.5822 & 0.6130 & 0.2496 \\
& +FreLLM4Rec & \textbf{0.6287} & \textbf{0.6311} & \textbf{0.3327} \\
\bottomrule
\end{tabular}
\end{table}

Table~\ref{tab:lora_performance} shows that LoRA fine-tuning yields mixed results across different backbones and datasets. While it provides notable improvements on LastFM (e.g., 11.6\% gain with Qwen2.5-7B), it shows inconsistent performance on other datasets, sometimes even underperforming the baseline (e.g., Llama-3.1-8B on All Beauty and Movies). In contrast, FreLLM4Rec consistently achieves substantial improvements across all configurations, demonstrating that explicit frequency-aware corrections provide a more robust and effective solution for preserving collaborative signals than implicit parameter adaptation through fine-tuning.

\subsection{Full-Ranking Evaluation}
\label{app:full_ranking}

To provide a more comprehensive evaluation, we conduct additional experiments under the full-ranking setting, where each model ranks the entire item catalog for every test sequence.

Table~\ref{tab:full_ranking} shows that FreLLM4Rec consistently outperforms all baselines across all datasets and metrics in the full-ranking setting.

\begin{table}[!t]
\centering
\caption{Performance comparison under full-ranking evaluation on the complete item catalog.}
\label{tab:full_ranking}
\resizebox{\columnwidth}{!}{%
\begin{tabular}{lcccccc}
\toprule
\multirow{2}{*}{Model} & \multicolumn{2}{c}{All Beauty} & \multicolumn{2}{c}{Luxury Beauty} & \multicolumn{2}{c}{LastFM} \\
\cmidrule(lr){2-3} \cmidrule(lr){4-5} \cmidrule(lr){6-7}
& N@10 & R@10 & N@10 & R@10 & N@10 & R@10 \\
\midrule
SASRec    & 0.4024 & 0.5266 & 0.2081 & 0.2883 & 0.0202 & 0.0396 \\
BERT4Rec  & 0.3552 & 0.5010 & 0.2111 & 0.2824 & 0.0249 & 0.0442 \\
MoRec     & 0.3819 & 0.5114 & 0.1731 & 0.2621 & 0.0164 & 0.0330 \\
FMLPRec   & 0.3917 & 0.5371 & 0.2001 & 0.2951 & 0.0243 & 0.0466 \\
BSARec    & 0.4240 & 0.5186 & 0.2309 & 0.3053 & 0.0239 & 0.0435 \\
SR-GNN    & 0.4134 & 0.4662 & 0.2087 & 0.2697 & 0.0141 & 0.0255 \\
MAERec    & 0.3985 & 0.5401 & 0.2283 & 0.3082 & 0.0182 & 0.0354 \\
LLM2Rec   & 0.3952 & 0.5258 & 0.2351 & 0.2883 & 0.0278 & 0.0486 \\
\midrule
\textbf{FreLLM4Rec} & \textbf{0.4451} & \textbf{0.5485} & \textbf{0.2529} & \textbf{0.3184} & \textbf{0.0317} & \textbf{0.0503} \\
\midrule
Improvement & +4.98\% & +1.56\% & +9.53\% & +4.29\% & +14.03\% & +3.50\% \\
\bottomrule
\end{tabular}%
}
\end{table}

\noindent\textbf{Full-Ranking Ablation Study.}
Table~\ref{tab:full_ranking_ablation} presents the ablation study under the full-ranking setting.
Both modules contribute consistently across all datasets.
Notably, removing ID embeddings causes more severe degradation under full-ranking than under the 100-item sampling setting, reflecting that collaborative signals are harder to compensate for with a larger candidate pool.
This confirms the robustness of our findings across evaluation protocols.

\begin{table}[t]
\centering
\caption{Ablation study under full-ranking evaluation (NDCG@10).}
\label{tab:full_ranking_ablation}
\begin{tabular}{lccc}
\toprule
Model Variant & All Beauty & Luxury Beauty & LastFM \\
\midrule
FreLLM4Rec (Full)  & \textbf{0.4451} & \textbf{0.2529} & \textbf{0.0317} \\
w/o TFM            & 0.4198          & 0.2391          & 0.0274          \\
w/o G-LPF          & 0.4231          & 0.2356          & 0.0285          \\
w/o G-LPF \& TFM   & 0.3981          & 0.2203          & 0.0224          \\
w/o Text emb       & 0.4083          & 0.2224          & 0.0196          \\
w/o ID emb         & 0.2876          & 0.1623          & 0.0071          \\
\bottomrule
\end{tabular}
\end{table}

\noindent\textbf{Full-Ranking Backbone Robustness.}
Table~\ref{tab:full_ranking_backbone} reports full-ranking results across different LLM backbones.
FreLLM4Rec consistently delivers 5--10\% NDCG@10 improvements across all backbones,
confirming that our approach addresses a fundamental architectural limitation rather than implementation-specific artifacts.

\begin{table}[t]
\centering
\caption{Full-ranking NDCG@10 across different LLM backbones. ``Base'' denotes the vanilla LLM baseline; ``+Fre'' denotes FreLLM4Rec.}
\label{tab:full_ranking_backbone}
\resizebox{\columnwidth}{!}{%
\begin{tabular}{lcccccc}
\toprule
\multirow{2}{*}{LLM Backbone}
  & \multicolumn{2}{c}{All Beauty}
  & \multicolumn{2}{c}{Luxury Beauty}
  & \multicolumn{2}{c}{LastFM} \\
\cmidrule(lr){2-3} \cmidrule(lr){4-5} \cmidrule(lr){6-7}
 & Base & +Fre & Base & +Fre & Base & +Fre \\
\midrule
Qwen2.5-7B      & 0.3981 & \textbf{0.4451} & 0.2203 & \textbf{0.2529} & 0.0224 & \textbf{0.0317} \\
Llama3.1-8B     & 0.3784 & \textbf{0.4112} & 0.2159 & \textbf{0.2364} & 0.0196 & \textbf{0.0268} \\
Mistral-7B-v0.3 & 0.3823 & \textbf{0.4067} & 0.2076 & \textbf{0.2408} & 0.0181 & \textbf{0.0253} \\
\bottomrule
\end{tabular}}
\end{table}

\begin{table}[!t]
\centering
\caption{NDCG@10 comparison isolating spectral modules from collaborative initialization. All variants use the same pre-trained ID embeddings.}
\label{tab:spectral_vs_smoothing}
\begin{tabular}{lccc}
\toprule
Variant & All Beauty & Luxury Beauty & LastFM \\
\midrule
ID only                  & 0.3741 & 0.2051 & 0.0146 \\
ID + Mean Smoothing      & 0.3749 & 0.2094 & 0.0153 \\
ID + G-LPF               & 0.3952 & 0.2201 & 0.0161 \\
ID only + TFM            & 0.4038 & 0.2198 & 0.0174 \\
\midrule
\textbf{FreLLM4Rec (Full)} & \textbf{0.4451} & \textbf{0.2529} & \textbf{0.0317} \\
\bottomrule
\end{tabular}
\end{table}

\subsection{Isolating Spectral Modules from Collaborative Initialization}
\label{app:spectral_vs_smoothing}

To disentangle the contribution of spectral frequency-domain design from the benefit of pre-trained ID embeddings, we introduce a non-spectral smoothing baseline: \textbf{ID + Mean Smoothing}, which replaces TFM with a simple moving-average operation $\tilde{\mathbf{H}}^{(l)}_t = \frac{1}{2w+1}\sum_{k=-w}^{w}\mathbf{H}^{(l)}_{t+k}, \quad w=1,$
applied after each Transformer layer while keeping the same pre-trained ID embeddings and G-LPF input purification. This isolates whether the gain comes from any smoothing operation or specifically from the frequency-domain Butterworth design.

Table~\ref{tab:spectral_vs_smoothing} reports NDCG@10 under this controlled comparison. Several observations are noteworthy. First, Mean Smoothing does outperform the ID-only baseline, confirming that temporal smoothing in general has a beneficial effect on collaborative signal preservation. Second, TFM consistently and substantially outperforms Mean Smoothing across all datasets, demonstrating that the frequency-selective Butterworth design—which attenuates high-frequency components while preserving the low-frequency passband without distortion—provides a distinct advantage over generic local averaging. Third, G-LPF alone also surpasses Mean Smoothing on most datasets, confirming that graph-aware input purification is a non-trivial contribution beyond simple smoothing. These results collectively confirm that the performance gains of FreLLM4Rec arise from principled frequency-domain design rather than from collaborative initialization alone.

\section{Related Work}
This section discusses related work from three aspects:

\noindent$\bullet$ \textbf{Sequential Recommendation Systems.}
Sequential recommendation learns item representations to predict items that users are likely to interact with in the future. From the item representation learning perspective, existing methods can be broadly categorized into three paradigms: ID-based methods, text-based methods, and hybrid ID-text methods.

ID-based methods assign each item a unique identifier and learn the corresponding representation using various sequence modeling techniques ~\cite{hidasi2015session, tang2018personalizedtopnsequentialrecommendation, kang2018self, sun2019bert4rec}. These ID-based representations primarily encode collaborative filtering signals by modeling multi-hop co-occurring patterns in sequential trajectories ~\cite{he2020lightgcn, wu2021self, xie2022contrastive, wu2019session}. Despite their effectiveness in capturing behavioral patterns, these methods cannot handle items or domains unseen during training, fundamentally limiting their generalization capability ~\cite{yuan2024cross, li2024llm}.
Text-based methods~\cite{liu2023pre, hou2023large} capture semantics but miss collaborative patterns such as item popularity and community preferences~\cite{ji2024genrec, lin2024rella}.

Hybrid ID-text methods attempt to combine both modalities through various fusion strategies: concatenating ID and text embeddings ~\cite{MoRec, yuan2025cilo}, using text to enhance ID representations ~\cite{wei2024llmrec, ren2024represenation}, or attention-based fusion mechanisms ~\cite{harte2024leveraging}. While these methods show improvements over single-modality approaches, they still require training on target domain data to learn effective representations. More importantly, they treat the fusion as a static combination without considering how these different signals interact and transform within deep neural architectures~\cite{he2025llm2reclargelanguagemodels}.

\begin{table}[!t]
\centering
\small
\caption{Performance (NDCG@10) comparison of FreLLM4Rec with different LLM backbones and parameter scales.}
\label{tab:llm_scales}
\begin{tabular}{lccc}
\toprule
LLM Backbone & All Beauty & Movies and TV & LastFM \\
\midrule
Llama3.2-1B & 0.5740 & 0.6180 & 0.2741 \\
Llama3.2-3B & 0.5799 & 0.6230 & 0.2843 \\
Llama3.1-8B & 0.5886 & 0.6402 & 0.3174 \\
Llama3-13B & 0.5974 & 0.6418 & 0.3225 \\
\midrule
Qwen2.5-0.5B & 0.6000 & 0.6113 & 0.2910 \\
Qwen2.5-7B & 0.6287 & 0.6311 & 0.3327 \\
\bottomrule
\end{tabular} 
\end{table}

\noindent$\bullet$ \textbf{Large Language Models in Recommendation.}
The application of large language models into recommendation systems represents a paradigm shift in how we approach personalized recommendation ~\cite{fan2023recommender, wu2023survey, li2024large}. This evolution has progressed through distinct phases, each addressing different aspects of the integration challenge. Initial approaches focused on leveraging semantic understanding through text-based methods, treating items as textual descriptions and framing recommendation as a language generation task ~\cite{geng2022recommendation, liu2023pre}. Subsequent developments introduced the embedding-as-token paradigm, enabling more efficient processing of user sequences while preserving structural information encoded in pre-trained embeddings~\cite{hou2024e4srec, zhang2024idgenrec, ji2024genrec}.

Recent studies show that semantic information captures item attributes and content features, but cannot fully model the collaborative patterns needed for effective recommendation~\cite{liao2024llara, wei2024llmrec, ren2024represenation}. This realization has led to hybrid approaches that combine semantic understanding with collaborative signals through various fusion strategies~\cite{sun2024large, he2025llm2reclargelanguagemodels, lin2024rella, wang2025oagir, yuan2025cilo, liu2025tmf, chen2025tler}. These developments underscore a critical insight: the most effective LLM-based recommender systems must successfully balance semantic understanding with collaborative information preservation. This balance presents fundamental challenges~\cite{zhang2024collm, sun2024large}: collaborative signals excel at capturing behavioral patterns for popular items, while semantic understanding handles sparse-data scenarios and enables cross-domain transfer~\cite{geng2022recommendation, bao2023tallrec}. Over-preserving collaborative signals harms generalization to cold-start items, whereas over-relying on semantics loses critical community preferences that cannot be inferred from content alone~\cite{he2020lightgcn, wang2019neural}. This tradeoff becomes particularly acute for LLM-based recommenders, as their pre-training on language tasks inherently biases them toward semantic correlations, making collaborative signal preservation more challenging than in traditional recommendation models~\cite{hou2024e4srec, liao2024llara}. Despite these advances, existing approaches predominantly treat LLMs as black boxes, focusing on input-output relationships without investigating how collaborative information evolves within the model's internal mechanisms.

\noindent$\bullet$ \textbf{Spectral Methods and Frequency Analysis in Recommendation.}
Graph signal processing reveals that collaborative information primarily resides in low-frequency components representing smooth, community-level patterns, while high-frequency components often encode noise~\cite{shuman2013emerging, ortega2018graph, he2020lightgcn, wu2019simplifying}. Effective GNNs inherently function as low-pass filters~\cite{nt2019revisiting}, preserving essential collaborative signals. Recent work incorporates frequency analysis into recommendation architectures~\cite{zhou2022filter, du2023frequency, shin2024attentiveinductivebiassequential} and knowledge distillation~\cite{zhu2025exploring} for architectural design purposes rather than as a diagnostic lens, leaving the internal spectral dynamics of LLM-based recommenders unexplored. Our work employs GSP as a diagnostic tool to reveal the previously unidentified Intra-Layer Spectral Attenuation phenomenon and proposes targeted spectral corrections to preserve collaborative information.

\section{Conclusion and Future Work}

\noindent\textbf{Conclusion.} In this paper, we identify the Intra-Layer Spectral Attenuation phenomenon in LLM-based recommendation systems, which systematically degrades collaborative signals essential for effective recommendation. Our spectral analysis reveals that LLMs progressively weaken low-frequency components that encode community preferences and interaction patterns, explaining why sophisticated language models often underperform in recommendation tasks. The proposed FreLLM4Rec approach introduces frequency-aware corrections through G-LPF and TFM modules that purify input signals and preserve spectral integrity throughout the network, successfully mitigating the spectral attenuation phenomenon.

\noindent\textbf{Limitation and future work.} While FreLLM4Rec demonstrates effectiveness and generalizability across diverse recommendation tasks, it still requires tuning of frequency-domain hyperparameters $\alpha$ and $\omega_c$. Developing adaptive frequency control mechanisms that automatically adjust filtering parameters based on data characteristics is reserved for future work. Additionally, exploring the spectral attenuation phenomenon in other domains where LLMs process structured information represents a potential research direction.

\bibliographystyle{ACM-Reference-Format}
\bibliography{main}

\appendix

\section{Wall-Clock Timing Analysis}
\label{app:timing_analysis}

To provide a comprehensive understanding of FreLLM4Rec's computational efficiency, we conduct detailed wall-clock timing measurements comparing training time per epoch and inference latency per user. We isolate the computational cost of each component by comparing three configurations: the vanilla LLM baseline, FreLLM4Rec without TFM (w/o TFM, which includes only G-LPF), and the complete FreLLM4Rec framework.

As shown in Table~\ref{tab:timing_analysis}, the G-LPF module operates as a one-time preprocessing step on item embeddings, requiring only 0.217 seconds on the LastFM dataset. Since this operation occurs before training and does not affect the forward pass, FreLLM4Rec (w/o TFM) exhibits identical training and inference times to the vanilla LLM baseline. The full FreLLM4Rec framework, which includes TFM modules after each Transformer layer, introduces a modest computational overhead of approximately 3.3\% in training time and 3.5\% in inference latency. This minimal increase is expected given that TFM's FFT operations have $\mathcal{O}(B \cdot d \cdot T \log T)$ complexity, which is asymptotically smaller than both the Transformer's self-attention ($\mathcal{O}(B \cdot T^2 \cdot d)$) and feed-forward network ($\mathcal{O}(B \cdot T \cdot d^2)$).

The cost-effectiveness of FreLLM4Rec becomes evident when considering the performance gains relative to computational overhead. While the overhead is minimal (3-4\%), the improvement in NDCG@10 reaches 48.8\% on LastFM, demonstrating that our frequency-aware corrections provide substantial value with negligible computational cost. This favorable trade-off makes FreLLM4Rec practical for production deployment.

\begin{table}[!t]
\centering
\caption{Wall-clock timing analysis on LastFM using Qwen2.5-7B backbone with batch size 32 on NVIDIA A800.}
\label{tab:timing_analysis}
\resizebox{\columnwidth}{!}{%
\begin{tabular}{lccc}
\toprule
Configuration & Training/Epoch & Inference/User & NDCG@10 \\
\midrule
Vanilla LLM & 6m 03s & 0.173s & 0.2236 \\
FreLLM4Rec (w/o TFM) & 6m 03s & 0.173s & 0.2507 \\
FreLLM4Rec (Full) & 6m 15s & 0.179s & 0.3327 \\
\midrule
Overhead & +3.3\% & +3.5\% & +48.8\% \\
\bottomrule
\end{tabular} }
\end{table}

\section{Connecting Temporal and Graph Frequency Domains}
\label{subsec:dft_gft_relationship}

\begin{figure*}[htbp]
    \centering
    \begin{subfigure}[b]{0.32\textwidth}
        \centering
        \includegraphics[width=\textwidth]{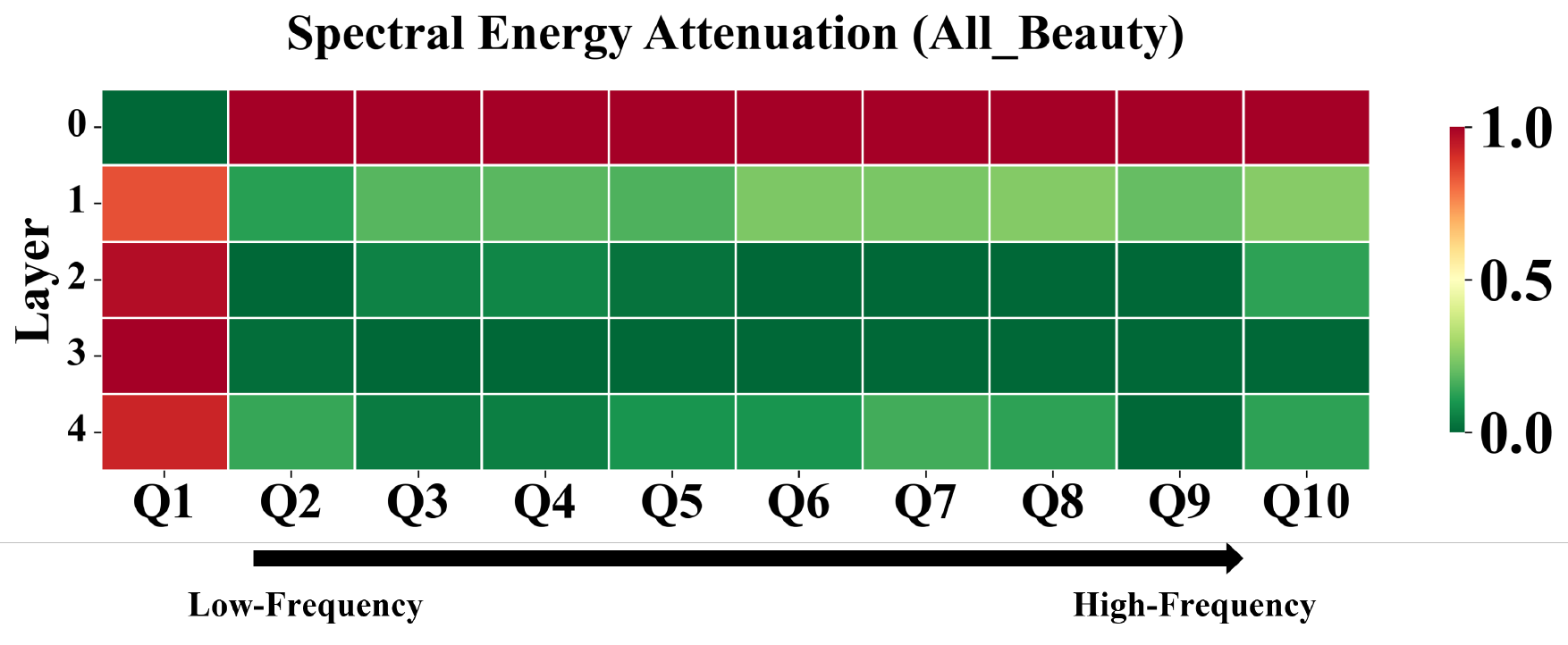}
        \caption{SASRec}
    \end{subfigure}
    \hfill
    \begin{subfigure}[b]{0.32\textwidth}
        \centering
        \includegraphics[width=\textwidth]{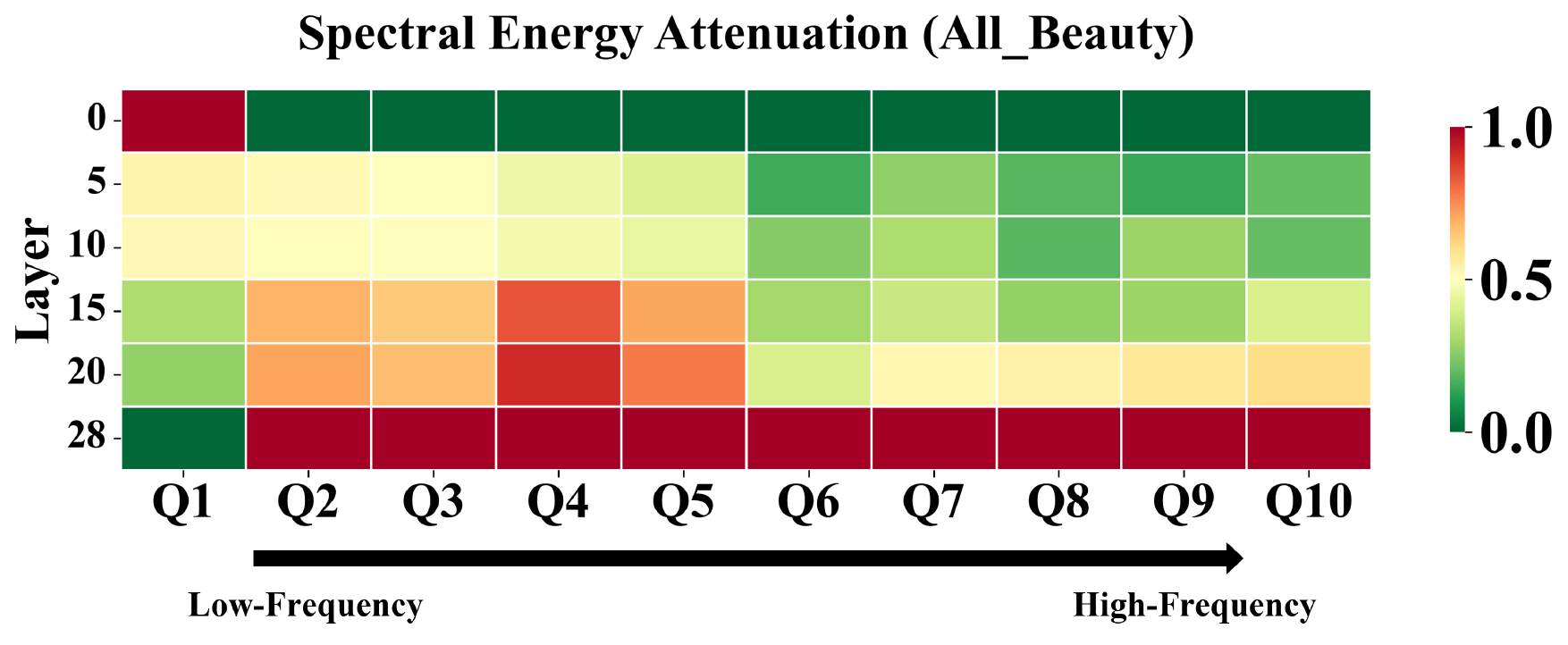}
        \caption{Vanilla LLM (Qwen2.5)}
    \end{subfigure}
    \hfill
    \begin{subfigure}[b]{0.32\textwidth}
        \centering
        \includegraphics[width=\textwidth]{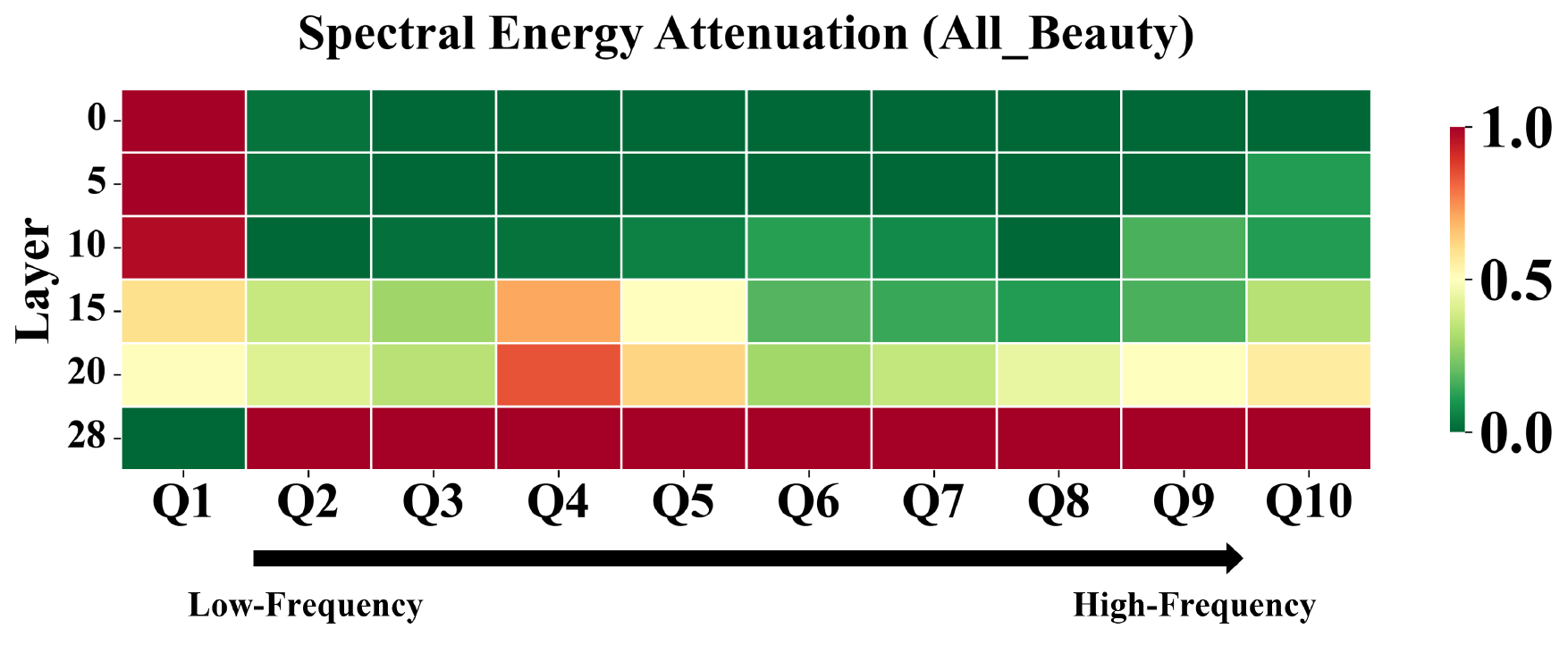}
        \caption{Vanilla LLM (Llama3.1)}
    \end{subfigure}
    \label{fig:all_beauty_decay}
    \centering
    \begin{subfigure}[b]{0.32\textwidth}
        \centering
        \includegraphics[width=\textwidth]{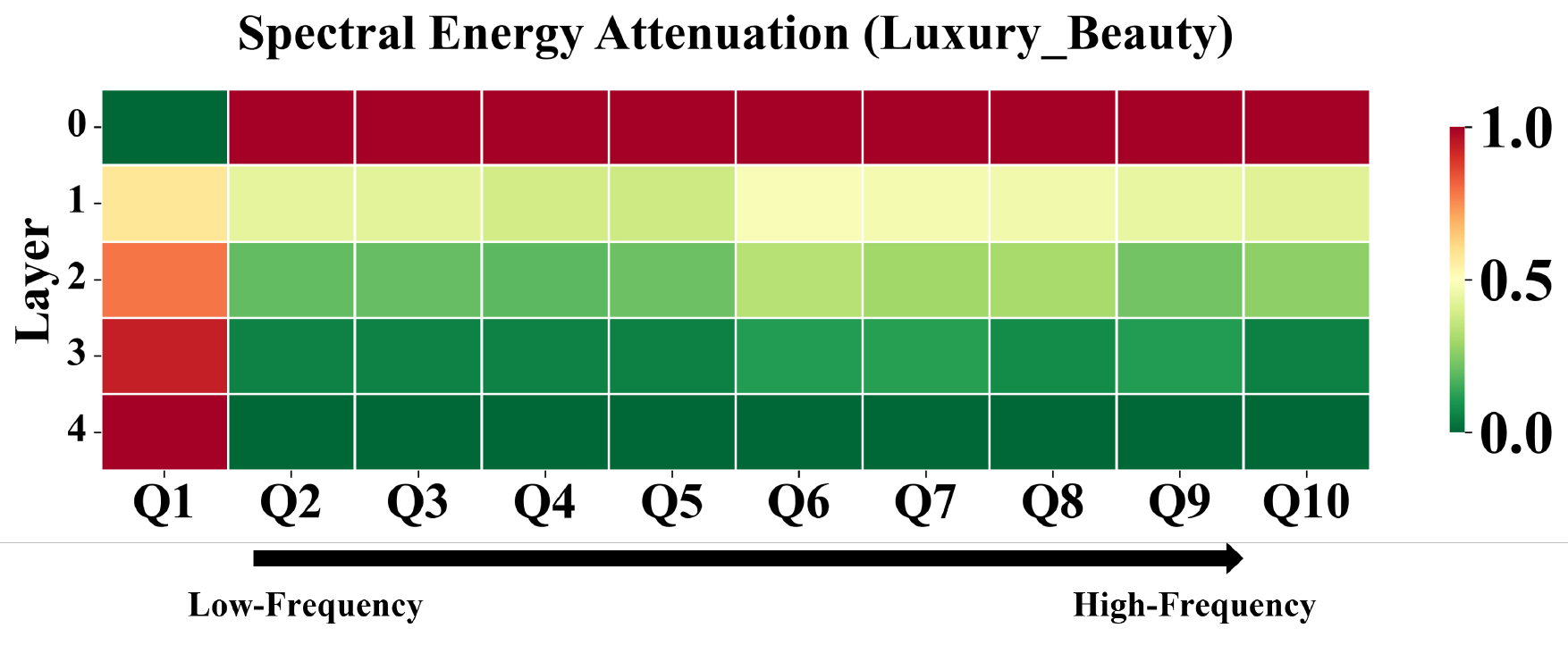}
        \caption{SASRec}
    \end{subfigure}
    \hfill
    \begin{subfigure}[b]{0.32\textwidth}
        \centering
        \includegraphics[width=\textwidth]{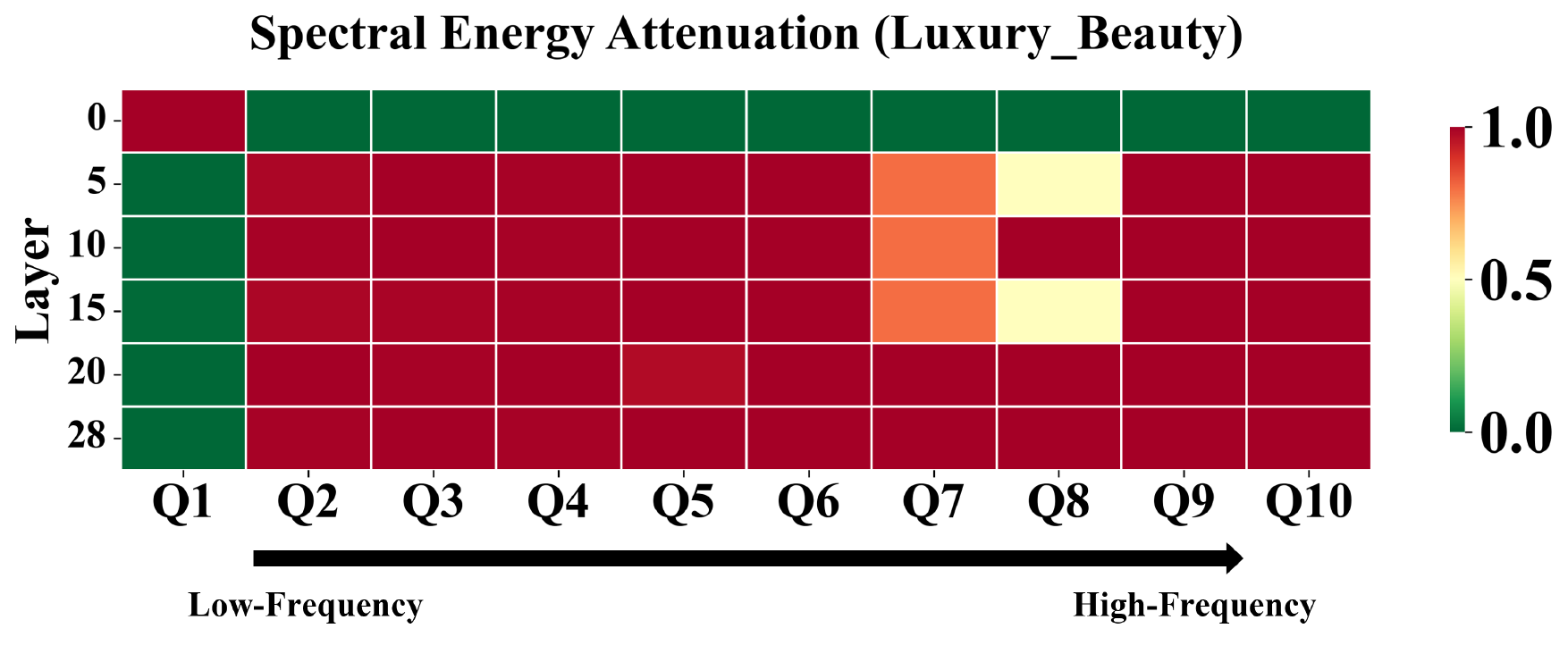}
        \caption{Vanilla LLM (Qwen2.5)}
    \end{subfigure}
    \hfill
    \begin{subfigure}[b]{0.32\textwidth}
        \centering
        \includegraphics[width=\textwidth]{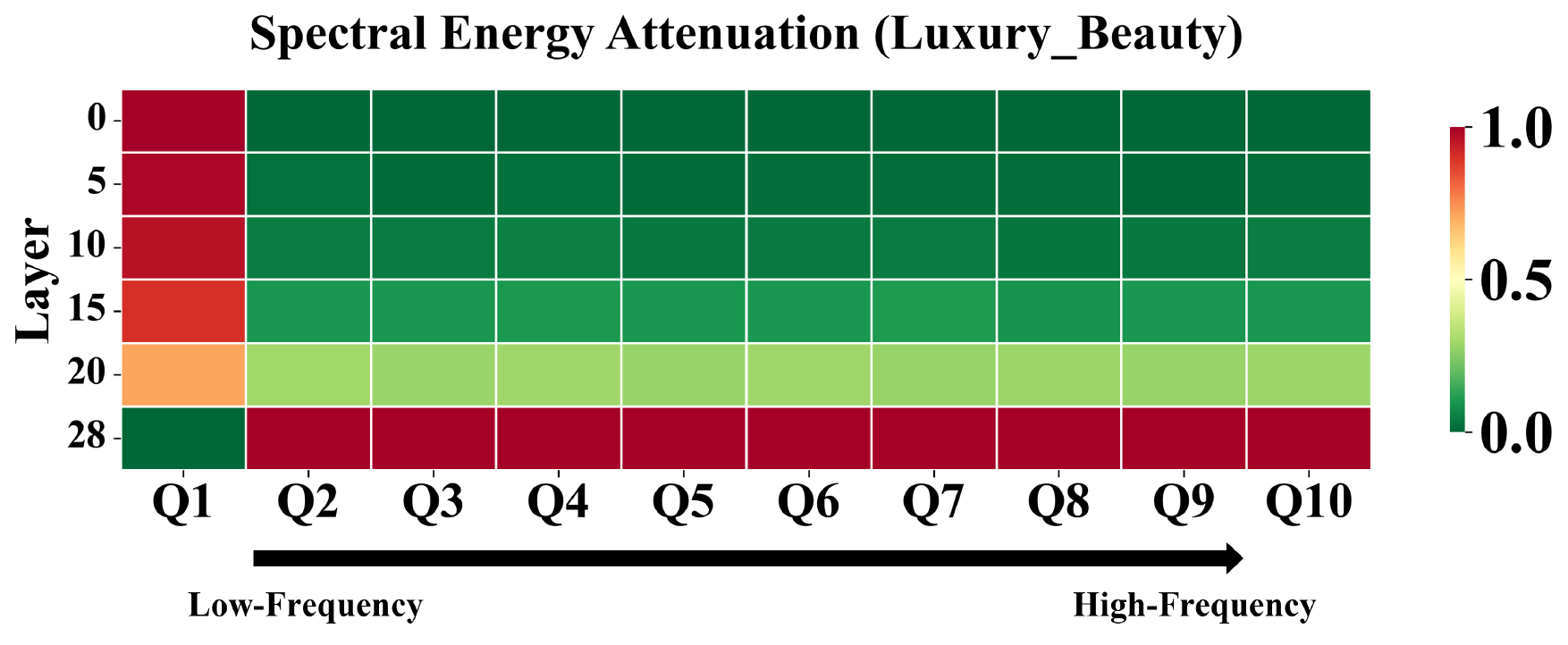}
        \caption{Vanilla LLM (Llama3.1)}
    \end{subfigure}
    \caption{Spectral attenuation phenomenon across different LLMs.
    Frequency-wise (column-wise) min-max normalization is applied to highlight relative cross-layer trends for each frequency component, making the evolution patterns more visible.}
    \label{fig:other_dataset_decay}
\end{figure*}

The Discrete Fourier Transform (DFT) for a temporal signal $\mathbf{x} = [x_0, x_1, \dots, x_{T-1}]^T$ is defined as:
\begin{equation}
\mathcal{F}(\mathbf{x})[k] = \sum_{t=0}^{T-1} x_t e^{-j2\pi kt/T},
\end{equation}
where $k = 0, 1, \dots, T-1$ represents the frequency index. In matrix form, this becomes $\hat{\mathbf{x}} = \mathbf{F} \mathbf{x}$, where the DFT matrix $\mathbf{F}$ has entries $F_{kn} = \frac{1}{\sqrt{T}} e^{-j2\pi kn/T}$.

The fundamental connection emerges when we recognize that the DFT can be interpreted as the Graph Fourier Transform (GFT) on a directed cyclic graph, commonly known as a ring graph. In this ring structure, each node connects to its temporal successor, creating a circular chain. The eigenvectors of this ring graph's Laplacian are precisely the DFT basis vectors~\cite{yu2023relationship}, establishing that temporal frequency analysis and graph spectral analysis are mathematically equivalent when applied to appropriately structured data.

This equivalence reveals that temporal frequency $k$ in the DFT corresponds to a graph eigenvalue:
\begin{equation}
\lambda_k = 2 - 2\cos(2\pi k/T).
\end{equation}
Low temporal frequencies (small $k$) correspond to small eigenvalues, representing slowly varying patterns. Conversely, high temporal frequencies (large $k$) correspond to large eigenvalues, capturing rapid fluctuations between adjacent elements.

For recommendation systems, this connection has profound implications. When we process a sequence of item embeddings representing a user's interaction history, temporal smoothing operations encourage adjacent items in the sequence to have similar representations. Under the reasonable assumption that users interact with similar items in temporal succession as a manifestation of preference locality~\cite{kang2018self}, temporal smoothing directly translates to enhanced smoothness on the Item co-occurrence graph.

\section{Proof of Theorem \ref{thm:temporal_to_graph}}
\label{app:proof_theorem}

\begin{proof}
The theorem states that applying a temporal low-pass filter to a signal enhances its graph smoothness, which in turn concentrates the signal's energy in the low-frequency bands of the graph spectrum. We prove this in two main parts.

\noindent\textbf{Part 1: Temporal low-pass filter enhances graph smoothness.}

Let $\mathbf{f} \in \mathbb{R}^T$ be a single-dimensional time series (a signal on the nodes of graph $\mathcal{G}$) and let $\mathbf{f}' = \text{LPF}_{\text{time}}(\mathbf{f})$ be the signal after applying a temporal low-pass filter. The graph smoothness of a signal is measured by the Laplacian quadratic form, $S_G(\mathbf{f}) = \mathbf{f}^T \mathbf{L} \mathbf{f}$, which can be expanded as:
\begin{equation}
S_G(\mathbf{f}) = \sum_{i=1}^T \sum_{j=1}^T w_{ij} (f_i - f_j)^2
\end{equation}
where $w_{ij}$ are the elements of the graph's adjacency matrix $\mathbf{W}$. Our goal is to show that $S_G(\mathbf{f}') \le S_G(\mathbf{f})$.

A temporal low-pass filter is a convolution operation that acts as a local averaging function. The value of the filtered signal at time $t$, $f'_t$, is a weighted average of the original signal's values in a local neighborhood around $t$. This inherently reduces the differences between adjacent points.

Under Assumption ~\ref{assump:locality} (Spatio-Temporal Locality), the adjacency weights $w_{ij}$ are large for nodes $i$ and $j$ that are close in the temporal sequence (i.e., $|i-j|$ is small). These terms with large $w_{ij}$ dominate the sum for $S_G(\mathbf{f})$. The local averaging effect of the temporal filter systematically reduces the difference $(f'_i - f'_j)^2$ for these dominant pairs where $|i-j|$ is small.

More formally, the difference $(f'_i - f'_j)^2$ in the filtered signal is a function of a set of local differences in the original signal. Let the low-pass filter be represented by a linear operator $\mathbf{H}$, so that $\mathbf{f}' = \mathbf{H}\mathbf{f}$. The difference becomes $(f'_i - f'_j)^2 = \left( \sum_k h_{i,k}f_k - \sum_k h_{j,k}f_k \right)^2$. Due to the smoothing nature of the convolution kernel $\mathbf{H}$, the resulting differences are attenuated. Invoking Jensen's Inequality ~\cite{boyd2004convex} for the convex function $\phi(x)=x^2$ provides a formal basis for this reduction in variance. Summing over all pairs $(i, j)$ weighted by $w_{ij}$, we conclude that the total graph smoothness decreases:
\begin{equation}
S_G(\mathbf{f}') = \sum_{i,j} w_{ij} (f'_i - f'_j)^2 \le \sum_{i,j} w_{ij} (f_i - f_j)^2 = S_G(\mathbf{f})
\end{equation}
This establishes the first part of the theorem.

\noindent\textbf{Part 2: Enhanced graph smoothness implies energy concentration in the graph low-frequency spectrum.}

From GSP, we have the identity that relates graph smoothness to the spectral domain:
\begin{equation}
S_G(\mathbf{f}) = \mathbf{f}^T \mathbf{L} \mathbf{f} = \hat{\mathbf{f}}^T \mathbf{\Lambda} \hat{\mathbf{f}} = \sum_{k=1}^T \lambda_k |\hat{f}_k|^2
\end{equation}
where $\hat{f}_k$ is the $k$-th GFT coefficient of $\mathbf{f}$ and $\lambda_k$ is the corresponding graph frequency (eigenvalue).

From Part 1, we have the inequality $S_G(\mathbf{f}') \le S_G(\mathbf{f})$. Applying the identity from above, we get:
\begin{equation}
\sum_{k=1}^T \lambda_k |\hat{f}'_k|^2 \le \sum_{k=1}^T \lambda_k |\hat{f}_k|^2
\label{eq:weighted_energy_inequality}
\end{equation}
This shows that the total energy, weighted by the graph frequencies, is reduced after filtering.

Furthermore, Parseval's theorem for the GFT states that the total energy of the signal is conserved in the spectral domain: $\|\mathbf{f}\|_2^2 = \sum_{k=1}^T |\hat{f}_k|^2$. While a temporal low-pass filter may slightly reduce the total signal energy, we can assume it is approximately conserved, or the signal can be re-normalized such that:
\begin{equation}
\sum_{k=1}^T |\hat{f}'_k|^2 \approx \sum_{k=1}^T |\hat{f}_k|^2
\label{eq:total_energy_conservation}
\end{equation}

We now have two conditions: the frequency-weighted sum of energy decreases (Eq. \ref{eq:weighted_energy_inequality}), while the total sum of energy remains approximately constant (Eq. \ref{eq:total_energy_conservation}). Given that the graph frequencies $\lambda_k$ are sorted in non-decreasing order ($0 \le \lambda_1 \le \dots \le \lambda_T$), the only way to satisfy both conditions simultaneously is for the energy distribution $|\hat{f}'_k|^2$ to shift its mass from terms with large $\lambda_k$ (high frequencies) to terms with small $\lambda_k$ (low frequencies).

Therefore, the application of a temporal low-pass filter leads to a redistribution of the signal's energy in the graph spectral domain.
This realizes more concentration on the low-frequency bands.
\end{proof}

\end{document}